%% file: paper.tex
\documentclass{article}

\usepackage{microtype}
\usepackage{graphicx}
\usepackage{subcaption}
\usepackage{booktabs} 

\usepackage{hyperref}

\usepackage[accepted]{icml2026}

\usepackage{amsmath}
\usepackage{amssymb}
\usepackage{mathtools}
\usepackage{amsthm}

\usepackage[capitalize,noabbrev]{cleveref}

\usepackage{comment}
\usepackage{color,soul}

\usepackage{url}

\usepackage{longtable}
\usepackage{soul}
\usepackage{multirow}
\usepackage{wrapfig}
\usepackage{svg}

\usepackage{microtype}
\usepackage{graphicx}
\usepackage{color,colortbl,tabularx}
\definecolor{Gray}{gray}{0.9}

\newcommand{\ourShort}{STEEL}
\newcommand{\ourLong}{Sample ThEn Evaluate Learner}

\usepackage{enumitem}

\theoremstyle{plain}

\theoremstyle{definition}

\theoremstyle{remark}

\usepackage[textsize=tiny]{todonotes}

\icmltitlerunning{Weight Space Learning for Certifiable Few-shot Transfer Learning}

\begin{document}

\twocolumn[
  \icmltitle{Weight Space Learning for Certifiable Few-shot Transfer Learning}



  \icmlsetsymbol{equal}{*}

  \begin{icmlauthorlist}
    \icmlauthor{Fady Rezk*}{samsung,edinburgh}
    \icmlauthor{Royson Lee}{samsung}
    \icmlauthor{Henry Gouk}{edinburgh}
    \icmlauthor{Timothy Hospedales}{samsung,edinburgh}
    \icmlauthor{Minyoung Kim}{samsung}
  \end{icmlauthorlist}

  \icmlaffiliation{samsung}{Samsung AI Center, Cambridge, UK}
  \icmlaffiliation{edinburgh}{School of Informatics, University of Edinburgh, UK}
  \icmlcorrespondingauthor{Fady Rezk}{s1985200@ed.ac.uk}

  \icmlkeywords{Machine Learning, ICML, Risk Certificates, AI Robustness, Trustworthy AI, Responsible AI, Performance Guarantees}

  \vskip 0.3in
]



\printAffiliationsAndNotice{*Work done while at Samsung AI Center, Cambridge UK.}  

\begin{abstract}
In contemporary deep learning, a prevalent and effective workflow for solving low-data problems is adapting powerful pre-trained foundation models (FMs) to new tasks via parameter-efficient fine-tuning (PEFT). However, while empirically effective, the resulting solutions lack generalisation guarantees to certify their accuracy - which may be required for ethical or legal reasons prior to deployment in high-importance applications. In this paper we develop a novel transfer learning approach that is designed to facilitate non-vacuous learning theoretic generalisation guarantees for downstream tasks, even in the low-shot regime. Specifically, we first use upstream tasks to train a {\em  distribution over PEFT parameters}. We then learn the downstream task by a {\em sample-and-evaluate} procedure -- sampling plausible PEFTs from the trained diffusion model and selecting the one with the highest likelihood on the downstream data. Crucially, this confines our model hypothesis to a {\em finite} set of PEFT samples. In contrast to the typical continuous hypothesis spaces of neural network weights, this facilitates tighter risk certificates. We instantiate our bound and show non-trivial generalization guarantees compared to existing learning approaches which lead to vacuous bounds in the low-shot regime. 
\end{abstract}

\section{Introduction}\label{sec:steel_into}
\begin{figure}[h]
    \centering
    \includegraphics[width=\columnwidth]{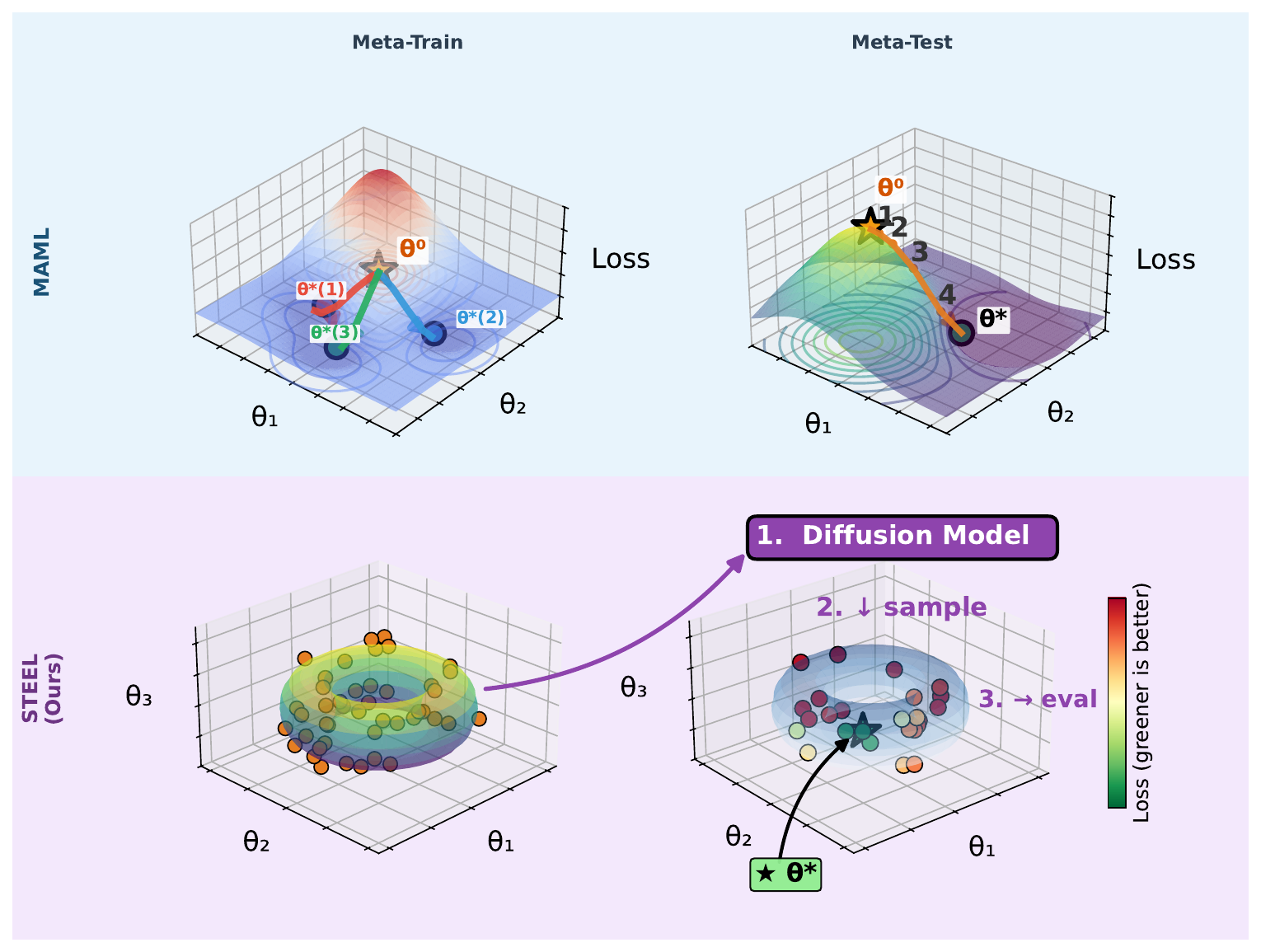}
    \caption{\textbf{Contrasting meta-learning paradigms.} \textit{Top row (MAML):} Traditional gradient-based meta-learning. During meta-training, MAML learns an initialization point $\theta^0$ positioned close to all task-specific optima. At meta-test, a few gradient descent steps adapt $\theta^0$ to the new task's optimum $\theta^*$. $\theta^0$ is the transferred meta-knowledge. \textit{Bottom row (\ourShort{}, Ours):} Manifold-based meta-learning. During meta-training, we collect adapters $\{\theta_i\}$ from upstream tasks and fit a diffusion model to capture their distribution---the task solution manifold. At meta-test, we sample candidates from the learned manifold and select $\theta^*$ by evaluating support set loss $r(\theta)$, replacing continuous optimization with discrete selection over a finite hypothesis space. The diffusion model is the transferred meta-knowledge.}
    \label{fig:steel_method_overview}
    \vspace{-2em}
\end{figure}

\begin{figure*}[h]
\centering
\includegraphics[width=\textwidth]{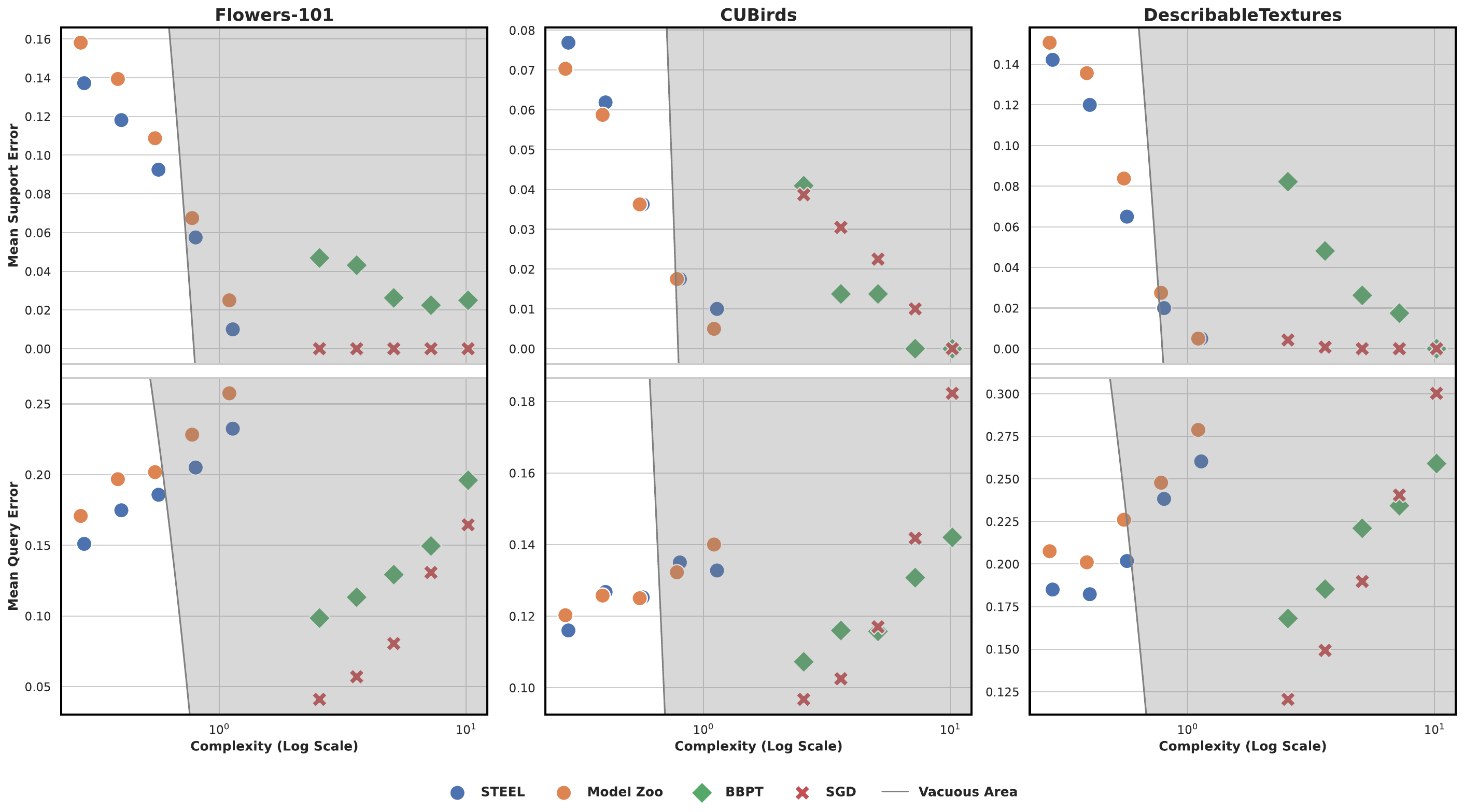}
\caption{Generalization bounds for adapting CLIP to novel tasks (5-way classification with 1–16 examples per class). Plots show classification error (y-axis) versus the complexity term (x-axis, log scale; square root terms from Equations \ref{eq:steel_bound}, and \ref{eq:qbound_best}). Top/Bottom: Mean support/query (train/test) error on new tasks. Shaded regions indicate vacuous bounds, where (support error + complexity) $\geq1$. Non-vacuous guarantees lie in the unshaded region. Competing methods (SGD, BBPT) fail to achieve non-vacuous bounds. In contrast, our method yields non-vacuous guarantees without significantly compromising training fit (top) or test accuracy (bottom).
}
\label{fig:steel_teaser_clip}
\vspace{-1em}
\end{figure*}

Generalisation certificates are crucial for high-importance applications where accuracy should be guaranteed for legal or ethical reasons. Guarantees should certify the minimum testing accuracy expected on unseen data drawn from the training distribution. However, it is hard to establish non-trivial guarantees for large neural networks, since  large learning capacity tends to produce looser guarantees. As such, there have only been a few successful demonstrations of non-vacuous guarantees for contemporary neural networks, even in the large-data regime \citep{dziugaite2017computing,ortiz2021tighter,lotfi2024nonvacuousLLM}. 

What about learning with sparse rather than large data? The problem of low-data learning is highly topical, due to the plethora of important limited-data applications \citep{wang2019fewShotSurvey}, but challenging due to the difficulty of learning a large number neural network parameters without overfitting. This need has inspired several lines of research that make use of different forms of knowledge transfer, including meta-learning \citep{hospedales20201metaSurveyPAMI} and parameter-efficient transfer learning (PEFT) \citep{hu2021lora} from foundation models. While PEFT methods have recently been more empirically effective, neither family of approach has produced methods that can provide low-shot generalisation guarantees, to our knowledge. From a learning theoretic perspective this is because existing algorithms still search a hypothesis space (e.g., all neural network weights $\theta\in\mathcal{R}^N$) large enough to make known bounds vacuous when instantiated. 

This paper introduces a novel approach to knowledge transfer that ultimately learns downstream tasks by \emph{picking from a finite set of hypothesis}, where the set of hypothesis is fit to the upstream tasks. Our method, \ourShort{} (\ourLong), facilitates using classic finite-hypothesis bounds, which are simple and tight, but not typically used in contemporary machine learning -- which focuses on learning continuous value neural network parameters. 

More specifically, in the upstream phase, we fit PEFT modules to available source tasks, and then train a parameter diffusion model to generate PEFTs according to this task distribution. In the downstream phase, we learn by sample-then-evaluate instead of traditional gradient descent. PEFT modules, unconditionally generated by the diffusion model, are scored using the target task training set, and learning is to choose the highest scoring module. Compared to the original set of upstream models, the diffusion model can be more compact, and can interpolate between the original model set to achieve higher accuracy. This procedure is gradient-free, which has some scalability benefits \citep{malladi2023finetuning,rezk2024liouna}, but more importantly it facilitates the use of PAC-Bayes finite-hypothesis bounds to provide non-vacuous guarantees, all while maintaining similar empirical accuracy to mainstream few-shot learning approaches. Figure~\ref{fig:steel_teaser_clip} shows some illustrative results, demonstrating our learner's ability to maintain practical efficacy while being constrained to low enough complexity to provide non-vacuous guarantees (white zone).

\textbf{Scope.} The novelty is methodological. The finite hypothesis class is the structural property that makes certification possible at all---no continuous-parameter method achieves this in the low-shot regime. This positions STEEL as a framework for \emph{certifiable} low-shot transfer: the right tool when deployment requires provable guarantees, complementary to (rather than replacing) unconstrained fine-tuning when raw accuracy is the sole objective. Empirically, STEEL is within a small margin of gradient-based fine-tuning on vision benchmarks (Table~\ref{tab:steel_vision_bound_stats}) and competitive on LLM benchmarks (Table~\ref{tab:steel_lamp_results}), while delivering certificates where every prior method is vacuous.

This opens a concrete research programme---spanning tightening the certificates within the STEEL framework, strengthening empirical accuracy, and extensions to cross-domain transfer. The rest of the paper develops this paradigm: Section~\ref{sec:steel_setup} reviews the risk-certificate and meta-learning background, Section~\ref{sec:steel_main} introduces STEEL and instantiates its PAC-Bayes bound, and Section~\ref{sec:steel_expmts} demonstrates non-vacuous certificates across LLM and vision benchmarks. Subsequently, we chart what comes next: the open research frontier this paradigm unlocks.

In summary, our contributions are: (1) Introducing a novel learning paradigm for gradient-free transfer learning designed to facilitate accuracy guarantees for downstream tasks, even in the low-shot regime with large foundational model. (2) The first practical demonstration of non-vacuous generalization bounds for low-shot learning in large language and vision architectures.

\section{Background}\label{sec:steel_setup}

\subsection{Risk Certificates for Deep Models}
Certifying model generalization performance is fundamental in theoretical machine learning~\citep{vapnik95,shalev2014understanding,fmm18}. Vapnik–Chervonenkis (VC), Rademacher, and PAC-Bayes bounds connect empirical risk (computable) to generalization risk (impossible to compute) through inequalities. Here we discuss the core concepts of risk certificates and why they are challenging to apply to deep models.

Let $h\in\mathcal{H}$ be a hypothesis (prediction function $y=h(x)$) and $\mathcal{H}$ the hypothesis space. In deep learning, $h$ corresponds to a model with parameters $\theta$, and $\Theta$ ($\ni\theta$) represents all possible parameter values, serving as the hypothesis space. A learning algorithm (i.e, SGD) selects $\theta$ from $\Theta$ given empirical data $S={(x_i,y_i)}_{i=1}^n$ sampled i.i.d. from distribution $T$.
The goal is minimizing generalization risk $R(\theta) = \mathbb{E}_{(x,y)\sim T}[l(\theta; x,y)]$, where $l(\theta; x,y)$ is the instance risk. Since computing $R(\theta)$ is impossible without access to $T$, we minimize empirical risk $r(\theta) = \frac{1}{n} \sum_{(x,y)\in S} l(\theta; x, y)$. Nevertheless, $r(\theta)$ is a surrogate for $R(\theta)$ and does not give any certificate about true generalization risk. Therefore, theoretical bounds relate $R$ and $r$ as follows. For any $\theta\in\Theta$:
\begin{equation}
R(\theta) \leq r(\theta) + \textrm{ComplexityTerm}(\dim(\Theta), n)
\label{eq:steel_generic_risk_bound}
\end{equation}
where the complexity term depends on data size $n$ and hypothesis space complexity $\dim(\Theta)$, decreasing as $n$ increases and $\dim({\Theta})$ decreases\footnote{PAC-Bayes bounds compute expected risk over a posterior on $\theta$ and include a complexity term from divergence to a prior. If we confine the posterior to concentrate at a point and use a flat prior, the form resembles (Eq.\ref{eq:steel_generic_risk_bound}). Some bounds relate $R$ and $r$ nonlinearly, but can be approximated by (Eq.\ref{eq:steel_generic_risk_bound}) without affecting our argument.}.
The right side of (\ref{eq:steel_generic_risk_bound}) is the risk certificate, which guarantees an upper bound on the risk of generalization. All terms on the right side are also computable. Different bounding methods, for example VC and Pac-Bayes bounds, produce different complexity terms. In totality, certificates above 1 (with $l\in[0,1]$) are vacuous, while we refer to those below 1 (or sometimes less than the random guess risk in the classification setting) as non-vacuous.

Traditional models like linear SVM achieve non-vacuous bounds through proper regularization~\citep{vapnik95,shalev2014understanding}.
However, conventional deep neural networks trained by gradient descent lack non-vacuous bounds unless $n$ is extremely large, due to high $\dim(\Theta)$ from numerous \textit{continuous} parameters. Even sparse adapter methods (e.g., LoRA~\citep{hu2021lora}, LoRA-XS~\citep{baazy2024loraxs}) face challenges from the continuous nature of $\Theta$. No existing deep learning approach achieves meaningful generalization bounds in low-shot settings.

\subsection{Meta-Learning}
We consider \emph{few-shot transfer learning}. In the upstream phase, we have training tasks $T_1,\dots,T_N$ sampled i.i.d.~from an unknown distribution over family of tasks $p_{\text{true}}(T)$, each with data to train a task-specific PEFT adapter. In the downstream phase, we receive a new task $T^* \sim p_{\text{true}}(T)$ observed only through a small support set $S^* = \{(x_i, y_i)\}_{i=1}^n$ with $n$ small (e.g., 8 or 16 examples). The goal is to adapt to $T^*$ using only $S^*$. This follows standard meta-learning assumptions~\citep{hospedales20201metaSurveyPAMI,maml}: downstream tasks are drawn from the same distribution as upstream tasks.

In standard meta-learning, the upstream phase learns meta-knowledge about task adaptation—such as a good initialization point~\citep{maml}, a metric space~\citep{protonet}, or an optimization procedure—that enables rapid adaptation to new tasks using only the small support set $S^*$. 

\textbf{The challenge.} Gradient-based adaptation on low-shot data $S^*$ leads to vacuous bounds due to high hypothesis space complexity. We need a hypothesis space $\Theta$ that is both (i) rich enough to contain good solutions for new tasks, and (ii) structured enough to yield non-vacuous bounds with small $n$ \cite{ravi2017optimization}.

\section{Proposed Approach}\label{sec:steel_main}
\subsection{High-Level Overview}\label{sec:steel_overview}

Figure~\ref{fig:steel_method_overview} contrasts standard meta-learning with our approach. Traditional methods (e.g., MAML~\citep{maml}) learn meta-knowledge such as good initialization points, then adapt via gradient descent on downstream tasks. In contrast, \ourShort{} learns the \emph{manifold of task solutions}—capturing where good adapters live in parameter space—then adapts by sampling and selecting from this learned distribution.

This paradigm shift enables two key advantages: (i) \textbf{Gradient-free adaptation}, avoiding the optimization on small downstream datasets that leads to overfitting, and (ii) \textbf{Finite hypothesis spaces}, making classical PAC-Bayes bounds applicable for the first time in few-shot deep learning.

\subsection{Method Details}\label{sec:steel_main_detail}

\textbf{Upstream: Learning the manifold over task solutions.} 
Given upstream task adapters $\overline{\Theta} = \{\theta_i\}_{i=1}^N$ where each $\theta_i$ solves task $T_i$, we treat these as samples from the true task distribution $p_{\text{true}}(T)$. Rather than meta-learning an adaptation procedure, we directly model the distribution of solutions via density estimation over $\{\theta_i\}_{i=1}^N$.

We construct the downstream hypothesis space $\Theta$ using one of two strategies:
\begin{itemize}[leftmargin=11pt, topsep=0pt, partopsep=0pt]
\item \textbf{Model zoo:} Set $\Theta = \overline{\Theta}$ directly. As $N$ increases, $\overline{\Theta}$ better approximates $p_{\text{true}}(T)$, but requires storing all $N$ adapters.

\item \textbf{Diffusion model (primary):} Train a diffusion model~\citep{ddpm} to learn $p(\theta)$ from $\overline{\Theta}$, then sample $\Theta$ from this model. This offers (i) \emph{compactness}—one generative model versus $N$ adapters, and (ii) \emph{interpolation}—generating novel adapters between training tasks, often improving coverage of $p_{\text{true}}(T)$. Recent work~\citep{nndiffu,protodiff,metadiff} demonstrates diffusion models effectively capture structure in neural network weight spaces.
\end{itemize}

\textbf{Downstream: Selection over finite hypotheses.}
Given test task support set $S^* = \{(x_i, y_i)\}_{i=1}^n$, few-shot adaptation is done by evaluate-then-select: :
\begin{align}
\theta^* = \arg\min_{\theta\in\Theta} 
\ r(\theta) = \frac{1}{n} \sum_{(x,y)\in S^*} l(\theta; x, y)
\label{eq:steel_strategy}
\end{align}

This is selection from a discrete set, not continuous optimization. The hypothesis space $\Theta$ encodes upstream inductive bias: each candidate is a plausible solution drawn from the learned task distribution.

However, the computational question naturally arises: {\em How do we solve (\ref{eq:steel_strategy}) efficiently?}
We consider two solutions:
\begin{itemize}
\item \textbf{Exhaustive search.}
Evaluate $r(\theta)$ for all $\theta \in \Theta$ and select the one with the lowest loss. This guarantees finding the optimal $\theta^*$, but is often computationally intractable (e.g., LLMs require a prohibitive number of forward passes or generations due to large $|\Theta|$).
\item \textbf{Hierarchical search.}
Use a tree-based approach such as hierarchical clustering of $\Theta$. Evaluate losses at the top level (e.g., cluster centroids or medoids), select the best cluster, and recurse within it. This reduces computation to $O(\log |\Theta|)$ and can yield a good approximation to $\theta^*$, but may result in suboptimal empirical loss $r(\theta)$ due to early pruning.
\end{itemize}

\subsection{Generalization Guarantees}\label{sec:steel_risk_cert}

The finite hypothesis space enables tight PAC-Bayes bounds. With a uniform prior over $\Theta$ and Dirac delta posterior at $\theta^*$, we obtain (Sec.~2.1.3 in~\citep{user_friendly}) that with probability at least $1-\epsilon$:
\begin{align}
R(\theta) \leq r(\theta) + C \cdot \sqrt{\frac{\log\frac{|\Theta|}{\epsilon}}{2n}} \quad \textrm{for any } \theta\in\Theta
\label{eq:steel_bound}
\end{align}
where $R(\theta) = \mathbb{E}_{(x,y)\sim T^*}[l(\theta; x,y)]$ is generalization risk and $C$ bounds loss ($0 \leq l \leq C$). $n$ is the total number of samples used and $|\Theta|$ is the total number of samples adapters from the diffusion model in our case.

The complexity term scales as $\sqrt{\log |\Theta|}$, so even $|\Theta| = 10^4$ maintains tight bounds. This makes our approach practical: a rich hypothesis space that represents the task distribution while yielding non-vacuous certificates. The bound is minimized by choosing $\theta^*$ with lowest empirical risk, directly justifying Eq.~\ref{eq:steel_strategy}.

\textbf{Computational considerations.}
For large $|\Theta|$, we use either \emph{exhaustive search} (evaluate all candidates; optimal but expensive) or \emph{hierarchical search} (cluster-based with $O(\log |\Theta|)$ complexity; faster but may select suboptimal $\theta$). The bound (Eq.~\ref{eq:steel_bound}) holds regardless; hierarchical search may yield higher $r(\theta)$, slightly loosening but never invalidating the certificate. See Appendix~\ref{app:steel_pseudo_code} for pseudo-code.

\subsection{Training Assumptions and Bounds Validity}\label{sec:bound_validity}
As in standard meta-learning, we assume upstream and downstream tasks are sampled i.i.d.\ from the same underlying task distribution. This ensures that the hypothesis space learned during the upstream phase contains high-quality solutions for unseen downstream tasks, a common assumption in few-shot transfer learning. We validate this empirically over a wide set of benchmarks demonstrating comparative downstream performance to SGD.

Importantly, the PAC-Bayes bound is valid without assuming downstream tasks match the upstream distribution. This assumption only affects tightness: distribution shift likely increases empirical risk, which loosens but doesn't invalidate the bound. Empirically, we find this is not an issue under mild distribution shifts. Crucially, even under distribution shift, the bound provides a principled criterion for deployment decisions—we can evaluate whether the certified risk is acceptable before deploying. This form of risk assessment is not available in gradient-based meta-learning, where adaptation always produces a solution without quantifying its reliability.

The bound's validity is also independent of the selection procedure used to choose $\theta^*$ from $\Theta$. Eq.~\ref{eq:steel_bound} holds for any $\theta \in \Theta$; exhaustive search, hierarchical search, or any heuristic (e.g., the two-stage procedure used for LaMP, Appendix~\ref{app:steel_lamp_hyperparams}) only affects \emph{which} adapter is returned and thus $r(\theta^*)$. A suboptimal selection loosens the certificate but never invalidates it, and the certificate is always computed on the finally deployed adapter.

\section{Related Work}\label{sec:steel_related}
\textbf{Risk certificates.}
While traditionally applied to simple models, recent work extends risk certificates to deep learning. Notable approaches include using data-dependent priors~\citep{ortiz2021tighter} and parameter quantization of PEFT adapters~\citep{lotfi2024nonvacuousLLM} in tabula-rasa and single source transfer learning respectively. Nevertheless, these still require large training sets to obtain non-vaucous bounds. In the meta-learning literature, ~\citet{meta_pac_bayes} proposed meta-learning PAC-Bayes bounds, which aims to facilitate tighter generalisation bounds by extracting a common prior from up-stream tasks in a multi-task setting. However, it was only demonstrated on toy problems and doesn't scale to large models due to the memory requirements of  nested gradient computations.

\textbf{Model diffusion methods.}
Several works explore diffusion for generating model parameters: NNDiffusion~\citep{nndiffu} for BN modules, ProtoDiff~\citep{protodiff} for ProtoNet-based few-shot learning, and MetaDiff~\citep{metadiff} and D2NWG~\citep{soro2025diffusionbased} for gradient-free meta-learning. Scaling properties of Diffusion based learning were also explored \citep{10.5555/3692070.3693860}. However, none provide risk certificates for the generated models. 

\textbf{Sparse adapter (PEFT) methods.}
Sparse adapters reduce learnable parameters, crucial for our diffusion-based sampling. While LoRA~\citep{hu2021lora} uses trainable low-rank matrices and VeRA~\citep{vera} uses fixed matrices with trainable diagonals, we adopt LoRA-XS~\citep{baazy2024loraxs}, which uses SVD with a trainable full matrix for the singular values. Such adapters have facilitated large-data guarantees \citep{lotfi2024nonvacuousLLM}, but in our framework they will facilitate low-shot guarantees.

\textbf{Connection to our Approach.} In terms of assumptions, we consider the same multi-task setting of meta-learning \citep{hospedales20201metaSurveyPAMI}, model-zoo \citep{huang2024lorahub} and model diffusion \citep{nndiffu} approaches. They all aim to facilitate downstream learning by knowledge transfer from upstream tasks. In terms of solution we share the benefit of model-zoo and model-diffusion methods in being able to use third-party off-the-shelf pre-trained upstream models, rather than requiring an expensive joint learning procedure like meta-learners. Like meta-learners and model-diffusers, but unlike zoo methods, we do learn a task-agnostic component (the diffusion model).

\section{Experiments}\label{sec:steel_expmts}
\textbf{Metrics for Certified Learning Quality:} In the low-shot learning context, we run many learning episodes with different random small training sets \citep{wang2019fewShotSurvey}. Thus we need evaluation metrics for the typical empirical performance and certificate strength instead of a single accuracy and certificate for one large scale learning \citep{lotfi2024nonvacuousLLM}. This is straightforward for empirical test performance: we report the average over episodes of the relevant task metric (e.g., accuracy, RMSE). For certificates, we report the minimum (best), median (typical) and maximum (worst) case error guarantee over all downstream tasks. Since many learners often produce vacuous certificates in the low-shot regime (e.g., guaranteed error not below 1), we use \emph{the proportion of tasks which have a non-vacuous certificate} as our leading metric. Finally, to quantify how tight the certificates are, we also report \emph{gap} - 
%
the average distance between certificates and query errors over all faithful bounds. More concretely, this is computed as $\sum_{i=0}^K\frac{\text{max}(0, \text{bound}_i - \text{query error}_i)}{K}$ where $K$ is the total number of downstream tasks.

\textbf{Competitors:} For single task methods that exploit only the foundation model and the target downstream task, we compare standard SGD, along with MeZO \citep{malladi2023finetuning} and BBPT \citep{bbpt} -- two state of the art gradient-free learners -- for fine-tuning.  For multi-task alternatives that exploit also the same task distribution as \ourShort{}, we compare the model zoo learner LoRA-Hub \citep{huang2024lorahub} and the meta-learner MetaPB \citep{meta_pac_bayes}. LoRA-Hub randomly samples from the model zoo and learns a new adapter as a linear combination of sampled adapters. 
Note that all methods are already combined with the same (experiment-specific) PEFT strategy for fair comparison. For SGD, MeZO, BBPT and MetaPB, this is an upgrade provided by us - they would definitely otherwise fail in the full parameter space. For SGD, MeZO, BBPT and LoRA-Hub, we combine them with recent quantization bound of \cite{lotfi2024nonvacuousLLM} (Eq.~\ref{eq:qbound_best}) to provide guarantees, MetaPB uses their own PAC Bayes bound, and \ourShort{} uses the bound in Eq.~\ref{eq:steel_bound}.
\begin{figure*}[t]
\begin{center}
\includegraphics[width=0.9\textwidth]{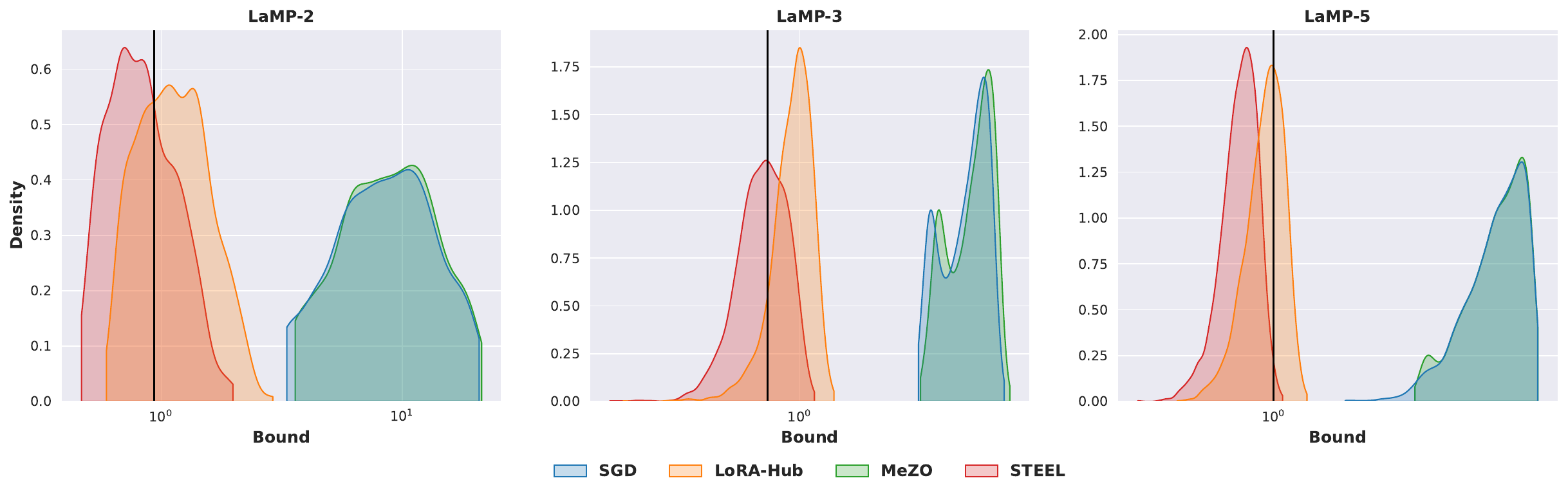}
\caption{Distribution of generalisation guarantees (x-axis, log scale) obtained over few-shot LLM adaptation episodes. Vertical lines indicate the vacuous bound threshold.  \ourShort{} provides a dramatically better distribution of provable generalisation outcomes compared to alternatives.}
\label{fig:steel_lamp_bound_distribution}
\end{center}
\vspace{-2em}
\end{figure*}

\subsection{Guarantees for few-shot LLM adaptation}\label{sec:steel_expmts_lamp}
\begin{table}[h]
    \centering
    \caption{LaMP LLM adaptation benchmark results.}
    \resizebox{\columnwidth}{!}{
        \begin{tabular}{ll>{\columncolor{Gray}}rrrr}
         \toprule
           & & \textbf{SGD} & \textbf{LoRA-Hub} & \textbf{MeZO} & \textbf{STEEL}\\
         \midrule
           \multirow{6}{*}{ \rotatebox[origin=c]{90}{\textbf{LaMP-2}} }
            &\% Non-Vacuous Tasks & 0.00\% & \underline{32.13\%} & 0.00\% & \textbf{65.12\%} \\
            &Average Gap & 9.03 & \underline{0.38} & 9.27 & \textbf{0.36} \\
            &Min Bound & 3.32 & \underline{0.59} & 3.60 & \textbf{0.47} \\
            &Median Bound & 8.52 & \underline{1.12} & 8.85 & \textbf{0.80} \\
            &Max Bound & 20.86 & \underline{2.90} & 21.36 & \textbf{1.99} \\
            &Accuracy$\uparrow$ & 63.25\% & 57.51\% & \underline{63.30\%} & \textbf{63.74\%} \\
            &F1$\uparrow$ & 56.15\% & 50.84\% & \underline{57.03\%} & \textbf{55.69\%} \\

            \midrule
            \multirow{8}{*}{ \rotatebox[origin=c]{90}{\textbf{LaMP-3}} }
            &\% Non-Vacuous Ratio & 0.00\% & \underline{10.88\%} & 0.00\% & \textbf{55.80\%} \\
            &Average Gap & 2.88 & \underline{0.31} & 2.95 & \textbf{0.24} \\
            &Min Bound & 2.26 & \underline{0.30} & 2.29 & \textbf{0.27} \\
            &Median Bound & 3.20 & \underline{0.97} & 3.32 & \textbf{0.78} \\
            &Max Bound & 4.07 & \underline{1.26} & 4.23 & \textbf{1.11} \\
            &MAE$\downarrow$ & 0.217 & \textbf{0.230} & 0.242 &\underline{0.231} \\
            &RMSE$\downarrow$ & 0.511 & \underline{0.526} & 0.531 & \textbf{0.524} \\
            & Expected Error$\downarrow$ & 0.244 & 0.649 & \textbf{0.297} & \underline{0.579} \\
        
            \midrule
            \multirow{6}{*}{ \rotatebox[origin=c]{90}{\textbf{LaMP-5}} }
            &\% Non-Vacuous Tasks & 0.00 & \underline{63.84} & 0.00 & \textbf{99.16} \\
            &Average Gap & 3.95 & \underline{0.38} & 3.96 & \textbf{0.28} \\
            &Min Bound & 1.61 & \underline{0.52} & 2.58 & \textbf{0.40} \\
            &Median Bound & 4.57 & \underline{0.96} & 4.57 & \textbf{0.81} \\
            &Max Bound & 5.86 & \underline{1.25} & 5.88 & \textbf{1.06} \\
            &ROUGE-1$\uparrow$ & 47.04\% & \underline{47.05\%} & 47.03\% & \textbf{47.22\%} \\
            &ROUGE-L$\uparrow$ & 42.79\% & \underline{42.75\%} & 42.73\% & \textbf{42.89\%} \\
         \bottomrule
    \end{tabular}
    }
    \label{tab:steel_lamp_results}
\end{table}
\textbf{Datasets:} We use the LaMP personalization benchmark \citep{salemi2024lamplargelanguagemodels} for low-shot LLM adaptation. LaMP consists of fixed training (seen) and evaluation (unseen) clients, each with support and query data. We evaluate on three datasets: LaMP-2 (nominal classification: movie tagging), LaMP-3 (ordinal classification: product rating), and LaMP-5 (text generation: scholarly title generation).

\textbf{Setup:} Following prior work \citep{tan-etal-2024-personalized,salemi2024lamplargelanguagemodels}, we fine-tune Flan-T5 base \citep{flant5} on all seen users' support data to build a task-specific base model. Then, we personalize it by training a LoRA-XS module per user \citep{baazy2024loraxs}, with rank 6 and alpha 16 (2592 tunable parameters). These modules, trained on the same support data used for base fine-tuning, form our model zoo. For efficiency, \ourShort{} uses hierarchical search (Section~\ref{sec:steel_setup}). Hyperparameters are in Appendix~\ref{app:steel_lamp_hyperparams}.
 
\paragraph{Results:} Our main contribution relates to the ability to provably certify the generalization of low-shot learning. In terms of low-shot LLM adaptation, Figure~\ref{fig:steel_lamp_bound_distribution} visualizes the distribution of certification outcomes over a large number of episodes for the three LaMP benchmarks. Taking note of the log-scale on the x-axis for generalization guarantee strength, we can see that our \ourShort{} learner provides dramatically better guarantees than conventional continuous-parameter learner alternatives, thanks to its discrete hypothesis space. The vertical lines indicate the threshold for vacuous bounds. Standard learners such as SGD and MeZO have no mass to the left of the threshold, while a substantial number of \ourShort{} learning episodes are non-vacuously guaranteed. Table~\ref{tab:steel_lamp_results} provides more detailed quantitative results in terms of various metrics. 
Notably, to assess provable generalization,  the data visualized in Figure~\ref{fig:steel_lamp_bound_distribution} is summarized as the \% non-vacuous metric (the fraction of episodes which have guarantees with a strength above chance-level), and the median guarantee strength across episodes.

From Table~\ref{tab:steel_lamp_results} results we can see that: (1) Standard supervised learning approaches such as (gradient-based) SGD and (gradient-free) MeZO have no non-vacuous episodes - no few-shot learning task can be guaranteed. (2) \ourShort{} has the most non-vacuous episodes for each benchmark, with almost every few-shot learning episode being guaranteed in the LaMP-5 benchmark. The median \ourShort{} episode also has a substantially non-vacuous guarantee for all three benchmarks. (3) Interestingly, LoRA-Hub combined with \citet{lotfi2024nonvacuousLLM}'s discretization bound also has some non-vacuous episodes, but less than \ourShort{}. (4) \ourShort{} has comparable empirical test performance to SGD and MeZO on LaMP-2 and LaMP-5, and comparable MAE and RMSE on LaMP-3, where its predictive distribution is less confident, which also limits certification there. Across all three, it provides a large improvement in certifiability.

\subsection{Guarantees for few-shot visual recognition}\label{sec:steel_expmts_clip}
\begin{figure*}[t]
\begin{center}
\includegraphics[width=\textwidth]{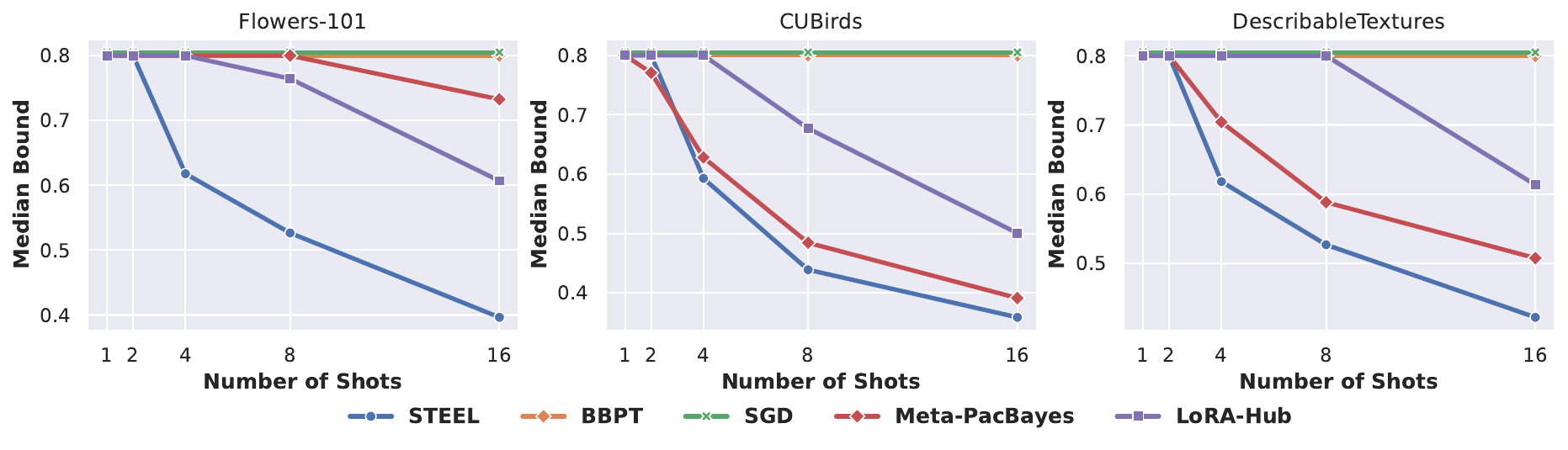}
\caption{Dependence of generalisation guarantee on training set size. Our finite-hypothesis class learner \ourShort{} achieves non-vacuous guarantees from 4-shot onward. Standard approaches provide no guarantees anywhere in this low-shot range.}
\label{fig:steel_clip_shots_vs_bound}
\end{center}
\end{figure*}

\begin{figure*}[t]
\begin{center}
\centerline{\includegraphics[width=\textwidth]{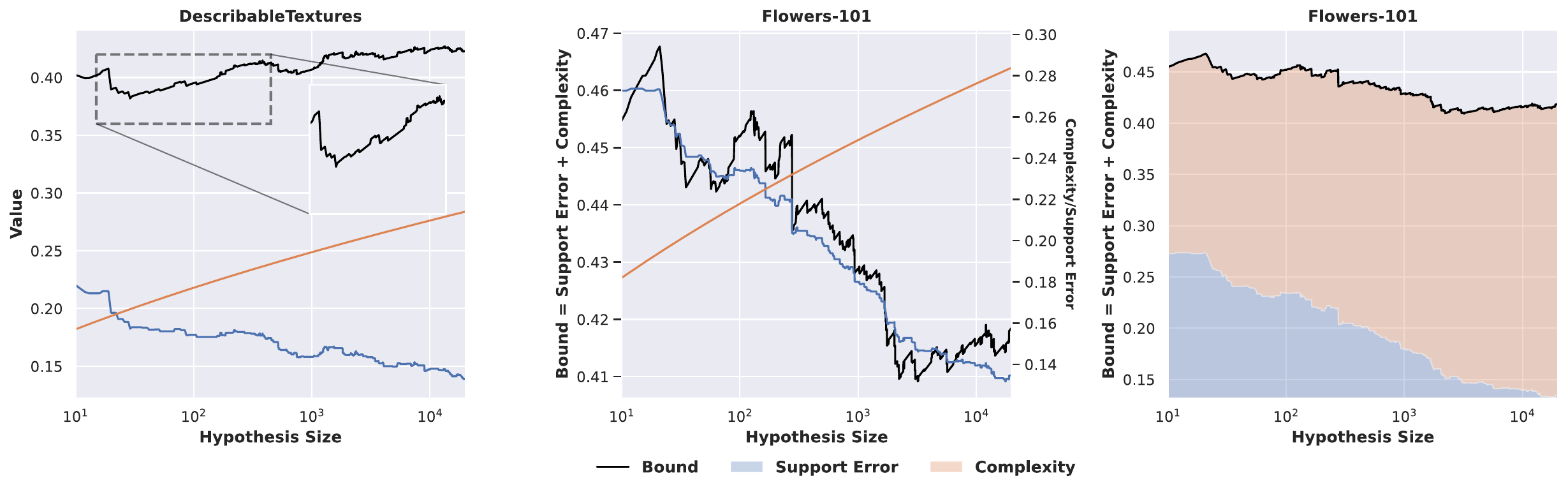}}
\caption{``Learning curves" illustrating empirical and certified learning dynamics of \ourShort{} with respect to samples/iterations, which is equivalent to hypothesis space size. More samples improves the training (support) error, while increasing the complexity penalty. The sum of these two terms instantiates the generalisation guarantee (Eq.~\ref{eq:steel_bound}) achieved for a given number of samples.}
\label{fig:steel_diffusion_samples_analysis}
\end{center}
\vspace{-1cm}
\end{figure*}

\textbf{Datasets:} We use fine-grained vision datasets with standard seen/unseen class splits: CUBirds \citep{cubirds}, FGVC-Aircraft \citep{aircrafts}, Describable Textures \citep{dtd}, and Flowers-101 \citep{flowers}. Splits follow  Meta-Dataset \citep{metadataset} for Flowers and Learn2Learn \citep{arnold2020learn2learnlibrarymetalearningresearch} for the others.

\textbf{Setup:} We sample $k$-shot, 5-way tasks following meta-learning conventions \citep{metadataset}, evaluating at $k \in \{1,2,4,8,16\}$. CLIP \citep{radford2021learningtransferablevisualmodels} serves as the backbone. For each task, we sample 5 classes, draw $n$ support examples per class, and adapt using CoOp \citep{zhou2022coop}—i.e., prompt tuning with a 2-token prefix per class (1024 total tunable parameters). The model zoo is built from training classes; evaluation uses disjoint unseen classes. Due to fast inference, we perform exhaustive search over sampled prompts. Meta-training, evaluation, and bound details for Meta-PB are provided in Appendix~\ref{appsec:pb_meta}. Hyperparameters for all other vision experiments are in Appendix~\ref{app:steel_vision_hyperparams}.

\paragraph{Results:} The results in terms of mean training and testing error versus complexity are summarized for three datasets in Figure~\ref{fig:steel_teaser_clip}. The dots for each learner reflect the training set sizes of 1, 2, 4, 8, and 16-examples per-class, and the white/grey zone separation delineates the space of non-vacuous vs vacuous bound outcomes. The main message is that, among the methods shown, only our finite hypothesis class approaches achieve any non-vacuous guarantees. Every result for the standard SGD and BBPT approaches is vacuous and cannot be guaranteed. For readability, the figure omits the multi-task baselines; Appendix~\ref{app:steel_full_complexity} shows it with LoRA-Hub and MetaPB included.

For the 16-shot case, these experiments are quantified in Table~\ref{tab:steel_vision_bound_stats}. Similarly to the results for LLM adaptation, we can see \ourShort{} matches or exceeds all baselines on the \% of non-vacuous episodes, and achieves the lowest min, median and max bound over episodes. Compared to the LLM case, \ourShort{}  pays a higher price in terms of empirical test accuracy compared to SGD. However, this is small compared to the stark difference in certification performance. The multi-task baselines (LoRA-Hub and MetaPB) certify most episodes non-vacuously on some datasets, but \ourShort{} is the most consistent, with the smallest drop in non-vacuous ratio on FGVCAircrafts ($97.50\%$), and offers the tightest certificates on every dataset. MetaPB as published was demonstrated with full parameter tuning on small-scale problems using many shots (MLP+MNIST). We devise a strong baseline from it by restricting it to PEFT adapters, instantiating its meta-training objective and re-deriving its test-time bound (Appendix~\ref{appsec:pb_meta}); with these adaptations it becomes non-vacuous, consistent with our principle that a constrained hypothesis space informed by upstream data enables certification.

\begin{table}[t]
    \centering
    \caption{Aggregate over 16-Shots 5-way visual recognition episodes. Standard Errors are reported in Appendix \ref{app:steel_vision_ste}}
    \label{tab:steel_vision_bound_stats}
    \small
    \resizebox{0.85\columnwidth}{!}{
    \begin{tabular}{l>{\columncolor{Gray}}rrrrr}
        \toprule
        Method & SGD & BBPT & LoRA-Hub & MetaPB & \ourShort \\
        \midrule
        \multicolumn{6}{c}{\textbf{CUBirds}}\\
        \midrule
        Non-Vacuous Ratio & 0.00\% & 0.00\% & 100.00\% & 100.00\% & 100.00\% \\
        Average Gap & 2.49 & 2.48 & 0.39 & 0.29 & \textbf{0.24} \\
        Min Bound & 2.55 & 2.55 & 0.44 & 0.31 & \textbf{0.30} \\
        Median Bound & 2.58 & 2.59 & 0.50 & 0.39 & \textbf{0.36} \\
        Max Bound & 2.64 & 2.68 & 0.61 & 0.54 & \textbf{0.47} \\
        Average Accuracy & 90.32\% & \textbf{89.27\%} & 87.82\% & 89.56\% & \underline{88.40\%} \\
        
        \midrule
        \multicolumn{6}{c}{\textbf{Describable Textures}}\\
        \midrule        
        Non-Vacuous Ratio & 0.00\% & 0.00\% & 100.00\% & 100.00\% & 100.00\% \\
        Average Gap & 2.43 & 2.46 & 0.39 & 0.31 & \textbf{0.24} \\
        Min Bound & 2.55 & 2.55 & 0.48 & 0.40 & \textbf{0.36} \\
        Median Bound & 2.55 & 2.63 & 0.61 & 0.51 & \textbf{0.42} \\
        Max Bound & 2.58 & 2.78 & 0.74 & 0.59 & \textbf{0.53} \\
        Average Accuracy & 87.95\% & \textbf{83.20\%} & 77.60\% & 80.15\% & \underline{81.50\%} \\

        \midrule
        \multicolumn{6}{c}{\textbf{FGVCAircrafts}}\\
        \midrule        
        Non-Vacuous Ratio & 0.00\% & 0.00\% & 37.50\% & 85.00\% & \textbf{97.50\%} \\
        Average Gap & 2.31 & 2.45 & 0.41 & 0.33 & \textbf{0.22} \\
        Min Bound & 2.58 & 2.64 & 0.59 & 0.58 & \textbf{0.45} \\
        Median Bound & 2.64 & 2.80 & 0.83 & 0.73 & \textbf{0.61} \\
        Max Bound & 2.78 & 3.01 & 1.05 & 0.95 & \textbf{0.85} \\
        Average Accuracy & 65.57\% & \textbf{62.37\%} & 57.70\% & 59.63\% & \underline{61.37\%} \\ 

        \midrule
        \multicolumn{6}{c}{\textbf{Flowers-102}}\\
        \midrule
        Non-Vacuous Ratio & 0.00\% & 0.00\% & 87.50\% & 65.00\% & \textbf{100.00\%} \\
        Average Gap & 2.51 & 2.50 & 0.41 & 0.49 & \textbf{0.27} \\
        Min Bound & 2.55 & 2.55 & 0.45 & 0.55 & \textbf{0.31} \\
        Median Bound & 2.55 & 2.59 & 0.61 & 0.73 & \textbf{0.40} \\
        Max Bound & 2.55 & 2.70 & 1.03 & 0.95 & \textbf{0.61} \\
        Average Accuracy & 95.90\% & \textbf{90.15\%} & 78.22\% & 72.02\% & \underline{84.90\%} \\
        \bottomrule
    \end{tabular}
    }
    \vspace{-2em}
\end{table}
Finally, Figure~\ref{fig:steel_clip_shots_vs_bound} highlights the evolution of the median generalization bound as a function of the training set size. For \ourShort{} it becomes non-vacuous from 4-shots onward, and the single-task approaches never become non-vacuous.\footnote{Note their bound is substantially worse than 0.8, but for simple visualization, we plot it as chance-level for 5-way classification.}

\subsection{Further Analysis}\label{sec:steel_further_analysis}
\textbf{Learning Curves:} We discuss and provide some insight into the learning process of our discrete hypothesis class learner. Standard gradient-descent takes repeated update steps to find a model that better fits a training set. By analogy, our gradient-free  \ourShort{} draws more  samples as it attempts to iteratively sample a model that better fits the training set. Our main experiments use a fixed number of 20,000 samples on all vision datasets. Figure~\ref{fig:steel_diffusion_samples_analysis} illustrates our learner's behaviour by showing the equivalent of a learning curve for our model. The x-axis is the number of samples drawn, and equivalently the learning theoretic hypothesis space size. Unlike SGD, this means that there is a direct dependence of hypothesis class complexity ($|\Theta|$ in Eq.~\ref{eq:steel_bound}) and the number of iterations/samples. This is reflected in the steadily increasing red complexity curve in Figure~\ref{fig:steel_diffusion_samples_analysis}(left, middle). We can also see that the training/support error goes down consistently over iterations/samples as the sampler progressively discovers better models. The generalisation bound (black line) is given by the sum of the training error and complexity. The figure illustrates one case (Flowers, middle, right) where the bound continues to improve up to a large number of samples/hypothesis size, because the continued improvement in training error outweighs the complexity gain. It also illustrates a case (DTD, left) where the training error improvement is slower and quite rapidly outweighed by the complexity gain, so that the best bound is actually achieved after quite a small number of samples.

\textbf{Sampler vs Zoo:} \ourShort{} compresses the upstream pre-trained models into a learned model generator. Selecting from the upstream models using downstream task performance as a criterion provides an alternative approach to learning that also corresponds to a finite hypothesis space. Our generator approach was motivated by ensuring scalability with respect to a large number of upstream models, and also to improve accuracy by enabling interpolation between upstream models rather than solely being limited to selecting one of them. Figure~\ref{fig:steel_teaser_clip} shows that \ourShort{}'s diffusion sampler tends to provide improved accuracy compared to its raw model zoo, especially for Flowers and DTD. On average across all datasets STEEL consistently outperforms Model Zoo. Detailed per-dataset per-shot performance is deferred to Appendix \ref{app:steel_extra_vision}, Table \ref{tab:steel_vision_accuracies}. Furthermore, we show in Appendix~\ref{app:certifying_zs} how to certify zero-shot CLIP using confidence intervals, and demonstrate that \ourShort{} not only achieves tighter guarantees but also substantially outperforms it in terms of empirical performance.

\begin{figure}{t}
\begin{center}
\includegraphics[width=1\linewidth]{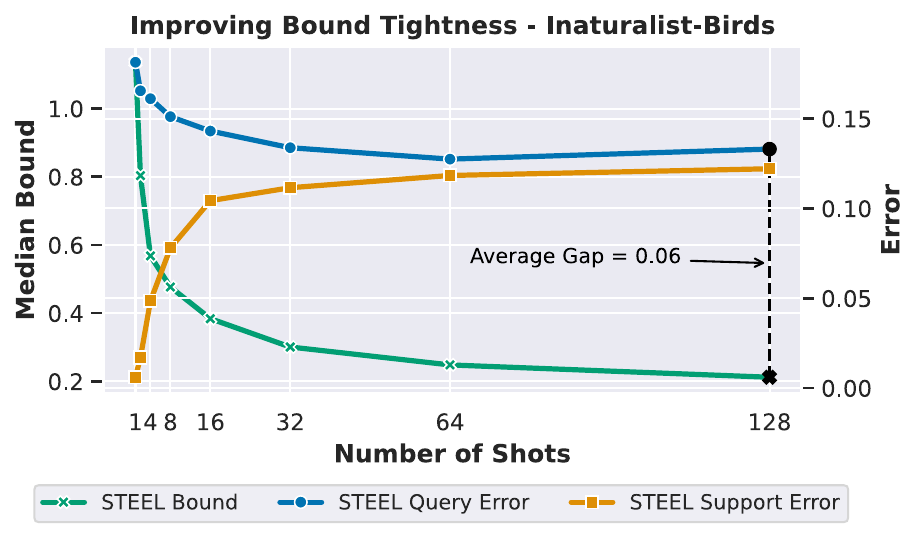}
\caption{\textbf{Support/query error and certified risk vs. support set size on iNaturalist birds.} As the number of shots increases, support and query errors converge, while the bound continues to tighten. The gap between the certified risk and empirical query error drops to 6\% at 128 shots, demonstrating the ability to produce tight certificates.}
\label{fig:steel_inaturalist}
\end{center}
\vspace{-2em}
\end{figure}

\textbf{Tightening STEEL Certificates:} So far, we have focused on the very low-shot regime to highlight our key result: the first non-vacuous certificate in this challenging setting. However, many applications may require stronger guarantees. To explore this, we increase the support set size from very low to moderately low shot using the iNaturalist dataset \citep{vanhorn2018inaturalistspeciesclassificationdetection} (birds subset, excluding overlapping classes), with a diffusion model pre-trained on CUB. This setup allows scaling up to 128-shot evaluation.
Figure~\ref{fig:steel_inaturalist} shows: (1) As the training set grows, the support error increases initially as there are more data points to fit, but soon stabilizes. The finite hypothesis-space does not problematically under-fit an increasingly large training set. (2) Query error decreases and saturates around 64 shots. (3) Crucially, the \ourShort{} bound keeps tightening: as support size grows, the complexity term continues to shrink even after support error plateaus, resulting in a certified risk just 6\% above empirical test error at 64–128 shots—while using far fewer examples than standard certification approaches \citep{lotfi2024nonvacuousLLM}. (4) Finally, it is important to note that this strong result is \emph{despite} the upstream-downstream distribution shift (CUB → iNaturalist), showing that \ourShort{} is robust to some task distribution shift.

\textbf{\ourShort{} Out-of-Distribution Robustness} Appendix~\ref{app:ood} provides a controlled study under graded in-domain corruption (CUB-C) and a head-to-head comparison against MAML, further substantiating STEEL's robustness under distribution shift.

\textbf{Diffusion Ablation:} Appendix~\ref{app:steel_diff_abl} presents ablation experiments comparing our diffusion model to alternative generative approaches. Results show that diffusion outperforms these baselines, supporting its use in our method.

\section{Discussion and Conclusion}\label{sec:discussion}
STEEL introduces a paradigm replacing continuous adaptation with discrete hypothesis selection: it trains a diffusion model on upstream task adapters to learn the ``solution manifold,'' then adapts to new tasks by sampling and selecting the best adapter based on support set loss. This shift from continuous to finite hypothesis spaces enables tight PAC-Bayes bounds, achieving the first non-vacuous generalization guarantees for few-shot learning with large models while maintaining comparable empirical accuracy. Across LLM personalisation and fine-grained vision benchmarks, STEEL delivers the highest fraction of non-vacuous certificates in the low-shot regime, often by a wide margin, while gradient-based, zeroth-order, and meta-PAC-Bayes baselines remain mostly vacuous at $n \leq 16$ shots. This establishes discrete hypothesis selection as a viable paradigm for certified foundation-model adaptation---not a theoretical curiosity, but a practical method.

The remainder of this section sketches the research programme this paradigm opens---tightening the certificate, closing the accuracy gap, and extending to cross-domain transfer---with full development in Appendix~\ref{sec:discussion}.

\textbf{Tightening the certificate.} Two directions promise tighter bounds without algorithmic change. The $\log|\Theta|$ term in Eq.~\ref{eq:steel_bound} treats every sampled adapter as contributing independently to complexity, but many produce near-identical predictions; replacing $|\Theta|$ with the prediction-space covering number $\mathcal{N}(\Theta, \varepsilon)$, computed via clustering on upstream reference data, decouples certificate strength from raw sample count. Separately, the diffusion model's own likelihood provides a natural data-dependent prior: substituting $-\log p_{\text{diff}}(\theta^*)$ for $\log|\Theta|$ tightens the bound whenever $\theta^*$ lies in a high-likelihood region---precisely when upstream and downstream distributions are aligned. Both repurpose information STEEL already computes (Appendix~\ref{app:future:tighten}).

\textbf{Closing the accuracy gap.} STEEL's certificate validity depends only on sampled adapters' empirical risk, not on the generator architecture or PEFT family. This makes the framework a passive beneficiary of progress in adjacent fields: stronger weight-space architectures such as SANE~\citep{10.5555/3692070.3693860} and improved PEFT modules translate directly to lower $r(\theta^*)$ and tighter bounds. Most distinctively, the discrete-hypothesis framework supports \emph{mixture hypothesis classes}: training separate diffusion models over different PEFT families (LoRA-XS, VeRA, prompt tuning) and constructing $\Theta$ as their union enables implicit per-task PEFT selection at only $\log K$ additional complexity cost. A complementary lever is \emph{sample-then-refine}: select $\theta^*$ via STEEL, then apply bounded continuous updates within an $\varepsilon$-ball, yielding a hybrid discrete--continuous procedure for which PAC-Bayes bounds with data-dependent priors remain applicable \citep{simpleshot}.

\textbf{Cross-domain transfer.} The cross-domain few-shot literature has converged on the view that robustness to domain shift is driven by the breadth of meta-training rather than by test-time adaptation~\citep{url,trim,eustratiadis2024neural}. STEEL is structurally well-suited to this regime: per-task adapters are trained independently before aggregation, so scaling to additional source domains amounts to extending the upstream pool---no episode balancing, no nested optimisation. Support-set-conditional generation pushes further by steering sampling toward downstream-relevant regions of weight space---the generative analogue of gradient-based adaptation, within a finite hypothesis class; the same mechanism also tightens empirical fit in-domain (Appendix~\ref{app:future:ood}).

\textbf{A new paradigm and what it opens.} Gradient-based meta-learning evolved incrementally over nearly a decade: from MAML~\citep{maml} on single-domain benchmarks (2017), to Meta-Dataset~\citep{metadataset} (2020), to methods designed for cross-domain transfer~\citep{url,trim}. STEEL introduces a different paradigm and establishes its feasibility together with the first non-vacuous certificates for few-shot deep learning. The directions above are the open frontier this paradigm unlocks. Establishing scaling laws for weight-space generative models would let practitioners predict the certificate--accuracy frontier as a function of upstream compute, hypothesis-space size, and downstream support-set size, and is one of the more exciting longer-horizon questions opened by this work.

\clearpage
\section*{Acknowledgements}
This project was supported by the Royal Academy of Engineering under the Research Fellowship programme.
\section*{Impact Statement}

This paper presents work whose goal is to advance the field of Machine Learning. There are many potential societal consequences of our work, and we mention a non-exhaustive concerns here. First, for LLM experiments, please note that the diffusion model is trained on adapters trained on user-specific data. We can use the diffusion model to interpolate between adapters to potentially avoid distributing the private model zoo among clients, but this does have potential for data leakage by memorization \cite{staab2024beyond}. 
It is worth investigating privacy preserving methods for such concerns \cite{miranda2024preservingprivacylargelanguage}.

On the vision side of our experiments, despite our benchmark being on fine-grained classification tasks of publicly available datasets, we necessitate the reminder of ethical concerns in computer vision as our method is easily applicable across domains and applications \cite{Waelen2023}.

\nocite{langley00}

\bibliography{paper}
\bibliographystyle{icml2026}

\newpage
\appendix
\onecolumn
\input{appendices}


\end{document}

%% file: appendices.tex
\appendix
\setcounter{secnumdepth}{2}
\renewcommand{\thesubsection}{\thesection.\arabic{subsection}}
\section{Existing Risk Bounds for Deep Models}\label{appsec:steel_other_bounds}

\subsection{Vanilla PAC-Bayes Bound}\label{appsec:vanilla_pb}

This is the vanilla, non-transfer learning bound.
As a baseline, one can also contrast with the vanilla PAC-Bayes bound (i.e., non-meta learning bound). This essentially follows the Cantoni's bound, and can be written as follows. With probability at least $1-\epsilon$, the following holds:
\begin{align}
\mathbb{E}_{Q(\theta)}[R(\theta)] \ \leq \ \mathbb{E}_{Q(\theta)}[r(\theta)] + \sqrt{\frac{\textrm{KL}(Q||\pi)+\log(1/\epsilon)}{2n}}
\label{eq:vanilla_pac_bayes}
\end{align}
Here $n$ is the test support size. 
We can set $\pi = \mathcal{N}(0, \kappa^2 I)$ for some fixed $\kappa_\pi$ and $Q(\theta) = \mathcal{N}(\mu, \Sigma^2)$ where the parameters $(\mu,\Sigma)$ can be learned by minimizing the right hand side. The sampled version $\theta^z = \mu + \sqrt{\Sigma}\cdot z$, $z\sim\mathcal{N}(0,I)$ can be used during the optimization. Once optimized, the minimum value of the right hand side serves as the error bound for the test task.

\subsection{Bound with Parameter-Level Quantization}\label{appsec:quantized_bound}

In~\citep{lotfi2024nonvacuousLLM}, they proposed a non-vacuous bound for the LLM based on the model parameter quantization (e.g., fixed-length floating point machine representation). There are several differences to our approach:

\begin{enumerate}
\item The paper is about LLM pre-training setup with large training data, and the bound would be vacuous if training data size is not large enough (e.g., $\geq$ 10K). 
\item They derive the same finite hypothesis space PAC-Bayes bound, but replace the $\log |H|$ term by $\log(1/p(h))$ where $p(h)$ is the prior likelihood, and $\log(1/p(h))$ is approximated and upper-bounded by $C(h)$ which is the number of bits for representing the hypothesis $h$.
\item The finite hypothesis space comes from the fixed-size floating point representation for real numbers (e.g., if there are $d$ trainable parameters, then $C(h) = d\cdot 32$), but to reduce it further, they propose what is called the SubLoRA, which is a random subspace representation (i.e., $\theta = P w$, $P =$ random subspace basis, $w$ = coefficients) of the LoRA $A$/$B$ matrices.
\item Also, instead of 32 bit for each of $d$ params, they do some clustering to reduce it to shorter coding, more precisely the arithmetic coding.
\end{enumerate}

The followings are some details of their bound derivation.
With probability at least $1-\epsilon$,
\begin{align}
R(\theta) \leq r(\theta) + C \cdot \sqrt{\frac{K(\theta) + 2\log K(\theta) + \log(1/\epsilon)}{2n}} 
\label{eq:qbound}
\end{align}
where $K(\theta)$ is the Kolmogorov complexity bound that can be estimated as:
\begin{align}
K(\theta) = \sum_{i=1}^d (\textrm{$\#$ of bits in the arithmetic coding of $\theta_i$})
\end{align}
where $d=\dim(\theta)$ for the PEFT parameters $\theta$. 
The arithmetic coding requires clustering of parameters $\theta_i$s, thus being dependent on the particular $\theta$ used. 
However, we can consider the best (i.e., the tightest) bound possible. That is, even if we have 1 bit for every $\theta_i$ (the minimal code length possible), $K(\theta)=d$, and plugging this into (\ref{eq:qbound}) yields:\footnote{Eq.~\ref{eq:qbound} omits the factor $\log 2$ that converts bits to nats; the complexity term of \citet{lotfi2024nonvacuousLLM} is $K(\theta)\log 2 + 2\log K(\theta)$. Our baseline numbers therefore correspond to about $1/\log 2 \approx 1.44$ bits per parameter rather than 1, still far below the 16 bits per parameter of the half-precision models we evaluate.}
\begin{align}
R(\theta) \leq r(\theta) + C \cdot \sqrt{\frac{d + 2\log d + \log(1/\epsilon)}{2n}} 
\label{eq:qbound_best}
\end{align}
which is the {\em best} scenario. Please note that this should be interpreted in the context of currently known bounds rather than all possible bounds.

In (\ref{eq:qbound}), as before, $R(\theta) = \mathbb{E}_{z\sim T^*}[l(\theta; z)]$ is the generalization error of $\theta$, 
$r(\theta) = \frac{1}{n} \sum_{z\in S^*} l(\theta; z)$ is the empirical error on the support data with size $n = |S^*|$, and $C$ is the maximal loss value (i.e., $0\leq l \leq C)$.

\subsection{PAC-Bayes Meta Learning Bound}\label{appsec:pb_meta}

In~\citep{meta_pac_bayes} they proposed an  extension of the PAC-Bayes bound for meta learning. Their meta learning algorithm aims to learn a distribution over the adaptation algorithms $A(S) \to Q(\theta)$ where $S$ is the support training data for a task and $Q(\theta)$ is the posterior distribution in the PAC-Bayes theorem. In their paper, any adaptation algorithm $A$ is allowable provided that $A$ internally makes use of a prior distributions in the PAC-Bayes theorem (e.g., the adaptation algorithms that try to minimize the PAC-Bayes bound), so that $A(S)$ becomes a function of the prior. And the meta learning model is a distribution $\rho$ that can sample the prior, and the meta learning amounts to learning the distribution $\rho$. 

Although they provided a concrete example in Appendix B.2 in their paper, the described meta training involves an optimization of a loss function that contains an optimal value of another problem, which is typically attained by gradient descent. Just like MAML-type algorithms with a large number of inner gradient steps, this renders the application of their theorem to large-scale model/data impractical. So we propose a more practical reptile-like adaptation strategy, essentially admitting a closed-form inner optimal solution, that is practical in large model/data scenarios. Note that under our setting the meta learning PAC-Bayes bound is still applicable since they said the theorem works for {\em any} adaptation algorithms.

\subsubsection{Concrete Cases with (Reptile-Like) Quadratic-Regularized Adaptation Algorithms}

Let $\theta$ be the parameters of a neural network or PEFT parameters. Denote $\dim(\theta)$ by $d$. 
The adaptation algorithm $A(S)$ returns a PAC-Bayes posterior $Q(\theta)=\mathcal{N}(\mu,\Sigma)$ for some task support data $S$ as input. We consider diagonal $\Sigma$. The algorithm $A$ itself is parametrized by the PAC-Bayes prior $P_1(\theta)=\mathcal{N}(\mu_1,\Sigma_1)$ since we only deal with those $A$s that internally use $P_1$ as a part of the algorithm without dependency on any other things. One candidate for $A$ is the algorithm that minimizes the PAC-Bayes bound with prior $P_1$ with respect to $Q(\theta)$. However, as said earlier, this incurs MAML-like optimization that is impractical.

Our choice of $A$ is as follows:

\textbf{(Step-1)} We first find $(\mu^{SGD},\Sigma^{SGD}) = \arg\min_{\mathcal{N}(\theta;\mu,\Sigma)} \mathbb{E}[l(\theta;S)]$ by gradient descent. Technically, this involves evaluating $l(\theta;S)$ with reparametrized $\theta=\mu+\sqrt{\Sigma}z$ for $z\sim\mathcal{N}(0,I)$, where the gradients take: $\nabla_\mu l(\theta;S) = \nabla_\theta l(\theta;S)$ and 
$\nabla_\Sigma l(\theta;S) = \nabla_\theta l(\theta;S) \cdot 0.5\Sigma^{-0.5}z$
by the chain rule. So it is as tractable as SGD since we only need conventional backprop $\nabla_\theta l(\theta;S)$ to get gradients with respect to $(\mu,\Sigma)$.

\textbf{(Step-2)} Solve the following quadratic optimization:
\begin{align}
(\mu^*,\Sigma^*) \ = \ \arg\min_{\mu,\Sigma} \ & ||(\mu,\Sigma) - (\mu^{SGD},\Sigma^{SGD})||^2 \ \notag\\
&+ \alpha \cdot ||(\mu,\Sigma) - (\mu_1,\Sigma_1)||^2
\end{align}
The intuition is that the adapted model $(\mu,\Sigma)$ has to be close to the fitted $(\mu^{SGD},\Sigma^{SGD})$ and also close to the prior $(\mu_1,\Sigma_1)$.
The trade-off coefficient $\alpha$ is user's choice, but when inspired by the PAC-Bayes bound objective, we can set $\alpha=1/\sqrt{|S|}$. 
The main benefit of this quadratic-regularized adaptation algorithm is that we have the closed-form solution:
\begin{align}
\mu^*(\mu_1, S) &= \frac{1}{1+\alpha} \mu^{SGD} + \frac{\alpha}{1+\alpha} \mu_1 \label{eq:mu_star} \\ 
\Sigma^*(\Sigma_1, S) &= \frac{1}{1+\alpha} \Sigma^{SGD} + \frac{\alpha}{1+\alpha} \Sigma_1
\label{eq:sigma_star}
\end{align}
Note that $(\mu^*,\Sigma^*)$ are functions of the prior $(\mu_1,\Sigma_1)$, and so we used the function notation in (\ref{eq:mu_star}--\ref{eq:sigma_star}) together with dependency on $S$ although $\mu^*$ does not depend on $\Sigma_1$ and $\Sigma^*$ similarly. We treat $(\mu^{SGD},\Sigma^{SGD})$ as constant.


Next we define the meta prior ($\pi$) and posterior ($\rho$) distributions for the meta learning PAC-Bayes bound. As suggested in~\citet{meta_pac_bayes},
\begin{align}
&\pi(A) = \mathcal{N}((\mu_1,\Sigma_1); 0, \kappa_\pi^2 I_{2d}) \\ 
&\rho(A) = \mathcal{N}((\mu_1,\Sigma_1); M_1, \kappa_\rho^2 I_{2d})
\end{align}
where $M_1 = (M^\mu_1, M^\Sigma_1)$ is the parameters for the meta posterior $\rho$. Here $\kappa_\pi$ and $\kappa_\rho$ are the fixed scales, and in their paper they used $\kappa_\pi = 100$ and $\kappa_\rho = 10^{-3}$.
Finally, they assume delta meta prior and posterior in the complexity term: $\mathbb{P}(A) = \mathbb{Q}(A) = \delta_{P_1}$.

This leads to the following meta training objective:
\begin{align}
&\min_{M_1} \ \hat{R}(\rho) \notag +\\
&\underbrace{\sqrt{\frac{\textrm{KL}(\rho||\pi)+\log(4\sqrt{N}/\epsilon)}{2N}}}_{=: T_1} \ +\\
&\ \underbrace{\sqrt{\frac{\textrm{KL}(\rho||\pi)+N\hat{E}(\rho)+\log(8nN/\epsilon)+1}{2nN}}}_{=: T_2}\notag
\end{align}

where

\begin{align}
\hat{R}(\rho) &= \mathbb{E}_{A\sim\rho}\Bigg[\frac{1}{N}\sum_{i=1}^N l(A(S_i); S_i)\Bigg] \notag\\
\hat{E}(\rho) &= \mathbb{E}_{A\sim\rho}\Bigg[\frac{1}{N}\sum_{i=1}^N\textrm{KL}(A(S_i)||P_1)\Bigg] \notag\\
\textrm{KL}(\rho||\pi) &= \frac{2d\kappa_\rho^2 + ||M_1||^2}{2\kappa_\pi^2} - d + 2d \log \frac{\kappa_\pi}{\kappa_\rho}\notag
\end{align}
Recall that $n$ is the support data size and $N$ is the number of meta training tasks.

\textbf{Meta training procedure (stochastic approximation version).} Since it is nearly infeasible to compute the objective for all $N$ tasks (for the purpose of one gradient update for $M_1$), we adopt the stochastic approximate optimization. That is, for each sampled task batch $B$ (out of $N$ tasks), we compute the followings (stochastic estimates of the objective and the gradient):
\begin{itemize}
\item $\hat{T}_0 \approx \hat{R}(\rho)$ where 
    \begin{align}
        &\hat{T}_0 = \frac{1}{|B|} \sum_{i \in B} l(\theta^{z_i}; S_i), \\ 
        &\theta^{z_i} = \mu^*(\mu_1,S_i) + \sqrt{\Sigma^*(\Sigma_1,S_i)} \cdot z_i, \ \ z_i \sim \mathcal{N}(0,I_d) \\
        & \mu_1 = M_1^\mu + \kappa_\rho \cdot z^\mu, \ \  z^\mu \sim \mathcal{N}(0,I_d)\\
        & \Sigma_1 = M_1^\Sigma + \kappa_\rho \cdot z^\Sigma, \ \ z^\Sigma \sim \mathcal{N}(0,I_d)
    \end{align}
Here $\mu^*(\mu_1,S_i)$ and $\Sigma^*(\mu_1,S_i)$ are determined by (\ref{eq:mu_star}--\ref{eq:sigma_star}). 
Note that $\hat{T}_0$ is a function of $M_1$, and the gradient of $\hat{T}_0$ with respect to $M_1$ can be obtained from $\nabla_{\theta^z}\hat{T}_0$ by the chain rule similarly as described previously. 
\item $T_1$ as it is since it is easy closed-form. 
\item $\hat{T}_2 \approx T_2$ where $\hat{T}_2$ only replaces $\hat{E}(\rho)$ by the following from $T_2$:
\begin{align}
\hat{E}(\rho) &\to \frac{1}{|B|} \sum_{i \in B} \textrm{KL}(A(S_i) \ || \ P_1) \\
&= \frac{1}{|B|} \sum_{i \in B} \textrm{KL}(\mathcal{N}(\mu^*, \Sigma^*) \ || \ \mathcal{N}(\mu_1, \Sigma_1))
\end{align}
This has a closed form from the Gaussian KL formula. 
\end{itemize}
As the above steps allow us to compute the meta training objective and its gradient, we can update $M_1$, and move on to a next batch.


\textbf{Test-time bound.} We certify each test task with a single-task PAC-Bayes bound whose prior is drawn from the meta-learned $\rho$. This bound does not use the meta-level theorem; meta-learning enters only through the choice of prior. We report only this per-task bound: the meta-level bound above is vacuous in our setting ($T_1 \approx 1.94$ from its $M_1$-independent part alone, with $d = 1024$, $\kappa_\pi/\kappa_\rho = 10^5$ and $N = 3{,}000$ meta-training tasks), so we use it only as the meta-training objective.

\emph{Prior.} We draw $P_1 = \mathcal{N}(M_1^\mu + \kappa_\rho z^\mu,\ M_1^\Sigma + \kappa_\rho z^\Sigma)$ (the variance is positive, since $M_1^\Sigma \ge 0.1$ and $z^\Sigma \ge 0$),
with $z^\mu, z^\Sigma$ drawn independently and uniformly from $[0,1)^d$, before observing the test support set $S^*$ with no selection among draws. Since $M_1$ and all hyperparameters are fixed using upstream tasks only, and $z^\mu, z^\Sigma$ are drawn independently of the data, $P_1$ is independent of $S^*$.

\emph{Posterior.} We fit $Q = \mathcal{N}(\mu, \Sigma)$, initialised at $P_1$, by minimising
\begin{equation}
\hat{L}_{\mathrm{CE}}(Q) + \sqrt{\frac{\mathrm{KL}(Q \,\|\, P_1) + \log(8n/\epsilon) + 1}{2n}}
\label{eq:lampert_test_time}
\end{equation}
with Adam, where $\hat{L}_{\mathrm{CE}}(Q)$ is the cross-entropy on $S^*$ of $\theta = \mu + \sqrt{\Sigma}\, z$, $z \sim \mathcal{N}(0, I_d)$. Cross-entropy serves only as a differentiable surrogate: because the bound below holds simultaneously for all posteriors $Q$, the posterior may depend on $S^*$ arbitrarily, and the choice of surrogate affects only tightness.

\emph{Certificate.} We use the 0-1 loss $l(\theta; x, y) = \mathbf{1}[h_\theta(x) \neq y] \in \{0, 1\}$. The certified quantity is the Gibbs risk $R(Q) = \mathbb{E}_{\theta \sim Q}[R(\theta)]$, and its empirical counterpart $\hat R(Q) = \mathbb{E}_{\theta \sim Q}[r(\theta)]$ has no closed form. We estimate it from $k$ i.i.d.\ draws $\theta_1, \dots, \theta_k \sim Q$ as $\hat R_k = \frac{1}{k} \sum_{j=1}^k r(\theta_j)$. Since each $r(\theta_j) \in [0, 1]$, with probability at least $1 - \delta'$ over the draws \citep{langford2001bounds, ortiz2021tighter},
\begin{equation}
\hat R(Q) \le \mathrm{kl}^{-1}\!\left(\hat R_k,\ \tfrac{\log(2/\delta')}{k}\right), \qquad \mathrm{kl}^{-1}(p, c) = \max\{q \in [p, 1] : \mathrm{kl}(p \,\|\, q) \le c\}.
\end{equation}
The complexity term involves no sampling, since $\mathrm{KL}(Q \,\|\, P_1)$ between diagonal Gaussians has a closed form; it only has to be evaluated at the reduced confidence level $\epsilon - \delta'$. Concretely, the single-task PAC-Bayes bound of \citet{maurer2004note} gives, with probability at least $1 - (\epsilon - \delta')$ over $S^*$,
\begin{equation}
\mathrm{kl}\big(\hat R(Q) \,\|\, R(Q)\big) \le \frac{\mathrm{KL}(Q \,\|\, P_1) + \log\big(\xi(n)/(\epsilon - \delta')\big)}{n}.
\end{equation}
We relax this in two ways: Pinsker's inequality $\mathrm{kl}(p \,\|\, q) \ge 2(p-q)^2$ turns it into an additive bound on $R(Q) - \hat R(Q)$, and since $\xi(n) \le n + 1$ for all $n \ge 1$, we replace $\log \xi(n)$ by the larger $\log(8n) + 1$. Substituting the Monte Carlo bound on $\hat R(Q)$ and combining the two events by a union bound gives, with probability at least $1 - \epsilon$ over $S^*$ and the draws,
\begin{equation}
R(Q) \le \mathrm{kl}^{-1}\!\left(\hat R_k,\ \tfrac{\log(2/\delta')}{k}\right) + \sqrt{\frac{\mathrm{KL}(Q \,\|\, P_1) + \log\!\big(8n/(\epsilon - \delta')\big) + 1}{2n}}.
\end{equation}
We use $\epsilon = 0.05$, $\delta' = 0.01$ and $k = 3{,}000$, and $n = |S^*|$ (five times the number of shots). The certificate applies to the Gibbs predictor, which draws a fresh $\theta \sim Q$ for each evaluation; the reported accuracy is one minus the mean 0-1 query error over $k = 3{,}000$ posterior draws, and the gap is computed from this query error as defined in Section~\ref{sec:steel_expmts}. The deterministic mean $\mu$ is not covered by this bound.

\textbf{Implementation note.} Our meta-training implementation departs from the objective above in several respects (the prior-sampling noise, the norm in the KL term, and the gradients of two terms). In effect, the prior variance reaches its lower clamp of $0.1$ early in meta-training and stays there, and the prior mean is fit to the upstream tasks by empirical risk minimisation. The learned prior is therefore less structured than the objective intends. This does not affect the validity of the test-time certificate, which requires only that $P_1$ be independent of $S^*$.


\section{Architecture and Training Recipes}\label{app:steel_hyperparams}
\subsection{Diffusion Architecture and Training Recipe}
First, the diffusion model forwad encoder uses a 1000 timesteps with a linear scheduler over noise between 1e-4 to 2e-2. For the decoder, we use an MLP network for the diffusion model with 3 hidden-layers. The hidden layer dimension is $4\times$ the size of it's input. This is 10,240 for LaMP (divisible by 512 for parallelization concerns) and 4096 for vision. A layer conditioned time embedding of the diffusion step is added (summed with) the hidden layer's hidden representation.

The time embedding is generated from the diffusion timestep using sinsusoidal embeddings as per the original DDPM model \citep{ddpm}. The dimensionality of the sinusoidal embedding is equal to the Diffusion MLP hidden dimension. Subsequently, the embedding is transformed using a two-layer MLP with first layer expanding the dimension to $4\times$ the network's hidden dimension and the second layer downscaling again to original hidden dimension. For example, on the vision experiments, the time embedding network has hidden dimension of $4098\times4$. Finally, to condition the time embedding computed by the two-layer MLP time embedding network for each layer, we apply a different linear transformation per diffusion hidden layer.

We train the diffusion model for 30K epochs for all experiments with a batch size of 1,024. We use the LAMB optimizer \citep{you2020largebatchoptimizationdeep} with a learning rate of 0.01. For vision experiments, we found that we can continue improving performance if we continue training for a second stage of 10K more epochs. For the second stage, we use a one-cycle learning rate scheduler \citep{smith2018superconvergencefasttrainingneural} with default hyperparameters (as found in pytorch). The maximum learning rate starts from 0.0004 reaching a maximum of 0.001 over a 1000 steps. We keep an exponential-moving average of the network weights throughout training with a decaying rate of 0.9999.

\subsection{Flan-T5 + LaMP Hyperparameters}\label{app:steel_lamp_hyperparams}
For training the base model across all datasets, we use LaMP's original recipe \citep{salemi2024lamplargelanguagemodels}. We use a batch size of 64, AdamW optimizer with a learning rate of 5e-5 and weight decay of size 0.0001. We use a linear warmup for the learning rate over 5\% of the total number of training steps.

To build the model zoo, we found that we required to tune Adam optimizer learning rate and per-dataset epochs per-dataset. We optimizer the hyperparameters to improve performance on the training split (seen users) query data. We use learning rates of 0.01, 0.01, 0.0001 and 20, 10, 10 epochs for LaMP-2, LaMP-3 and LaMP-5 respectively. For all datasets, we used a linear warmup for the learning rate over 5\% of the total number of training steps. We used the same recipe to train a per-user LoRA-XS model on the unseen users/novel tasks.

For LoRA-Hub, we used default hyperparameters as proposed by the original authors. First, we sample 20 random adapters from the model zoo. The weights of the linear combination is initialized with zeros and truncate min/max weights to -1.5/1.5. We do maximum inference steps of 40 with Nevergrad default hyperparameters.

Finally, we transform the SGD number of epochs to MeZO. The authors used 32$\times$ as many epochs as SGD in the original paper \citep{malladi2023finetuning}. This roughly corresponds to 320 epochs of finetuning for meta-training each task. We tested the three learning rates proposed to search over by the authors. We fixed a learning rate of 1e-3 across all datasets because we consistently found 1e-4 to not learn and 1e-2 to be unstable.

For LaMP, we sample 10K LoRA-XS modules from the diffusion model. We use k-means clustering on the diffusion samples to produce $N_c$ clusters where $N_c$ is chosen as the minimum Silhouette score. We evaluate N between [2,150] inclusive. For each cluster, we find the medoid; the adapter closest to the centroid of the cluster. During evaluation, we choose a cluster and evaluate all adapters therein. From the cluster, we short-list the best 15 adapters using the Flan-T5 training loss. On the best 15 adapters, we use text generation to produce an answer with greedy sampling. Using the LaMP benchmark proposed model selection metric for each dataset, we select the ``winning" adapter.

\subsubsection{Bound Metrics}
For support errors, we use 1-Accuracy for LaMP-2, and 1-ROUGE-1 for LaMP-5. For LaMP-3, both RMSE and MAE are not bounded. Therefore, we use the expected error for the dataset. First, we convert the ordinal vectors to one-hot encodings. Subsequently, we calculate the absolute error between the one-hot label and the predicted probabilities (softmax over the five rating tokens), divided by 2. This equals one minus the probability of the true rating. This guarantees the error to be bounded in the [0-1] range inclusive. We use this metric as support error term.

\textbf{LaMP-2 Intricacies:} LaMP benchmark has one query sample per user. For LaMP-3 and LaMP-5, this suffices since the generated support error is continuous. Nevertheless, for LaMP-2, the accuracy term, which we use to derive the support error, becomes the 0/1 loss. Therefore, we split the support data in novel tasks to support and query data with ratio 80\% and 20\% respectively. If the split generates only 1 query samples, we move one sample from support to query to have a minimum of two-samples in query. We use the same split for all evaluations across SGD, MeZO, LoRA-Hub and our proposed methods. For reproducibility, all splits were done deterministically. Furthermore, we did not constrain the split to have same classes across both support and query. LaMP classification tasks are long-tailed. Therefore, for a novel task, a user might have classes X and Y in support but the query ends up with classes A and B making it a more challenging benchmark for all methods.

Finally, we truncate the support sizes of LaMP-3 and LaMP-5 to 256 samples across all methods. This is done deterministically for reproducibility. The reason for truncating the dataset is pure computational concerns.

\subsubsection{Model Zoo Size}
For LaMP, we build a model zoo by training one PEFT adapter per-task in the dataset. Each user is treated as one task. This yields 3820, 20,000, and 9,682 total adapters/tasks in LaMP-2, LaMP-3, and LaMP-5 respectively.

\subsubsection{Compute Resources}
All experiments were conducted on 10xA40 GPUs. The base model finetuning for generating task specific models took a couple of hours with a batch size of 64 using 2 GPUs. Time to train an adapter for one user using one GPUs varies from dataset to dataset and from user to user depending on their available dataset size. This takes as low as 10 seconds and up to 2 mins per user. Diffusion model training on the collected model zoo varied among datasets too since each epoch was of different total number of steps because of different dataset sizes. We used one A40 GPU per diffusion model and train using the recipe described earlier.

Meanwhile, downstream adaptation using \ourShort{} varies between datasets. For LaMP-2, this averaged 0.8 minutes per user (50 A40 GPU hours total across all users). For comparison, SGD adaptation takes 10 seconds to 2 minutes per user.
Similarly, on LaMP-2 MeZO takes approximately 2.5$\times$ the wall-clock time of SGD, since it runs 16$\times$ more epochs (320 vs.\ 20) while each step is about 7$\times$ faster. This makes \ourShort{} faster than MeZO on LaMP-2. LaMP-5 is slower, averaging 7.6 minutes per user (315 A40 GPU hours in total), because it requires text generation.

\subsection{CLIP + CoOP Hyperparameters}\label{app:steel_vision_hyperparams}
To build the model zoo, we used the authors original hyperparameters to train CoOP. This is SGD with a learning rate of 0.002. For Flowers-102, we train for 200 epochs. For DTD, FGVCAircraft and CUBirds, we trained for up to 300 steps and found a One Cycle learning rate useful to stabilize training. These same hyperparameters were used to evaluate SGD on novel tasks. For BBPT, we use the authors default hyperparameters \citep{bbpt}. We found that the method converges within 8000 ``API call". We attempted to run for a budget of 20K as our diffusion model offers but found that performance did not improve. Please note that the original authors reduce the dimensionality of the prompt using a small network because evolutionary optimization struggles in high-dimension. They reduce dimensionality to 512. Nevertheless, since we train only 2-tokens (dimensionality=1,024), then we do not use the small dimensionality reduction network.

Finally, for MetaPB, we train the upstream model for 20 epochs on a random subset of 3{,}000 adapters from our model zoo which takes 24 hours similarly to our diffusion model training. We use Adam with default learning rate of 0.001 for both upstream and downstream. Downstream adaptation was done for 300 epochs with same Adam optimizer setting. We train both means and variances. Variances are clipped to 0.1 for numerical stability. Epochs and meta-training dataset size were chosen on meta-training distribution qualitatively to achieve healthy training curves (meta-training loss is lowering and converging within 24 hours; the same GPU cost we give to STEEL’s diffusion model training). 

For vision experiments, we found that exhaustive search was fast enough even though we sample 20K adapters from the diffusion model.

\textbf{Bound Metrics:} we use $1-\text{Accuracy}$ as the support error term in our bound calculation for all vision experiments except for MetaPB whose support term is the Monte-Carlo-corrected Gibbs error as explained in Appendix~\ref{appsec:pb_meta}.

\textbf{Model Zoo Size:} We randomly sample 16-shot 5-way tasks for building the model zoo from each respective dataset. We train 10,000 total tasks per-dataset for the model zoo. The diffusion model is trained on this model zoo. Once trained, we sample the diffusion model once and fix the samples across all downstream evaluation for 1, 2, 4, 8 and 16 shots.

\textbf{Compute Resources:} For SGD, training one task upstream for model zoo collection or downstream for gradient descent adaptation takes 10-15 seconds. BBPT downstream takes 10-15 seconds as well. Training the diffusion model on a model zoo, since we fix model zoo size in CLIP experiments, was the same among all experiments. Each diffusion model took 24 A40 GPU hours to train. \ourShort{} takes 30 second to 1 minute downstream for adapting to a new task. Finally, meta-training phase of MetaPB takes 24 hours similarly to our diffusion model training. MetaPB downstream takes 15 second per task too.

\section{Extra Vision Results}\label{app:steel_extra_vision}
\subsection{Certifying Zero-Shot Performance}\label{app:certifying_zs}
\begin{table}[t]
\centering
\caption{Comparing median bounds between \ourShort and zero-shot CLIP given a 16-shots support set.}
\begin{tabular}{lllll}
\toprule
\textbf{Median Bound} & \textbf{Flowers} & \textbf{CUB} & \textbf{DTD} & \textbf{FGVVCA} \\
\midrule
STEEL & 0.40 (+0.03) & 0.36 (-0.06) & 0.42 (+0.11) & 0.61 (+0.10) \\
ZS CLIP & 0.43 & 0.30 & 0.53 & 0.71 \\
\bottomrule
\end{tabular}
\end{table}
One might assume that zero-shot CLIP is a strong \textit{certified} baseline because it is evaluated without any fine-tuning, and the support and query sets are drawn i.i.d. from the same distribution. This can lead to the impression that its support-set accuracy reflects test-time performance. However, these are still empirical estimates and do not account for uncertainty due to sampling. In contrast, our certificates explicitly bound the generalization gap from limited support data. To facilitate a fairer comparison, we compute a Langford-style test-set bound (Theorem 3.3 by \citet{JMLR:v6:langford05a}) using zero-shot CLIP's support accuracy from Table 3, treating it as a proxy for true performance. The resulting guarantees are included in Table \ref{app:certifying_zs}. Although zero-shot CLIP performs well empirically, our STEEL method achieves tighter guarantees and often stronger accuracy in the few-shot setting.

\subsection{Effect of Candidate Pool Size}
We conduct an ablation to test whether \ourShort{}'s gains come from the diffusion model's learned distribution or simply from evaluating more candidate adapters in vision tasks. To investigate this, we compare three conditions: selecting from the 10K model zoo directly, generating and evaluating 10K samples from \ourShort{}, and generating and evaluating 20K samples from \ourShort{}. Results on 16-shot vision tasks are shown in Table~\ref{tab:pool_size}.

\begin{table}[t]
\centering
\begin{tabular}{lccc}
\toprule
Dataset & Zoo (10K) & \ourShort{} (10K) & \ourShort{} (20K) \\
\midrule
CUBirds & 87.97 & \underline{88.30} & \textbf{88.40} \\
DescribableTextures & 79.25 & \underline{81.42} & \textbf{81.50} \\
FGVCAircrafts & 61.02 & \textbf{61.50} & \underline{61.37} \\
Flowers-102 & 82.92 & \underline{84.22} & \textbf{84.90} \\
\midrule
Average & 77.79 & \underline{78.86} & \textbf{79.04} \\
\bottomrule
\end{tabular}
\caption{Comparison of candidate pool size and source. \ourShort{} at 10K samples already outperforms selecting from the 10K zoo, indicating gains come from the learned distribution rather than pool size. Increasing to 20K provides marginal additional improvement.}
\label{tab:pool_size}
\end{table}

\subsection{Accuracy Results Across Shots}
Results are provided in table \ref{tab:steel_vision_accuracies}.
\begin{table}[t]
    \centering
    \caption{CLIP+CoOp few-shot learning. Accuracies over different number of shots.}
    \begin{tabular}{lc>{\columncolor{Gray}}cccccc}
        \toprule
        \textbf{Dataset} & \textbf{Zero Shot} & \textbf{SGD} & \textbf{BBPT} & \textbf{Lora-Hub} & \textbf{Meta-PB} & \textbf{Model Zoo} & \textbf{\ourShort{}} \\
        \midrule
        \multicolumn{8}{c}{1-Shots}\\
        \midrule
        CUBirds & 83.82\% & 81.77\% & 85.80\% & 86.45\% & 88.25\% & 86.00\% & 86.72\% \\
        DescribableTextures & 67.29\% & 69.97\% & 74.10\% & 69.75\% & 78.55\% & 72.12\% & 73.97\% \\
        FGVCAircraft & 47.44\% & 53.00\% & 55.05\% & 54.05\% & 56.46\% & 54.37\% & 55.40\% \\
        Flowers-101 & 81.14\% & 83.55\% & 80.40\% & 73.02\% & 54.20\% & 74.25\% & 76.75\% \\
        Avg & 69.92\% & 72.07\% & 73.84\% & 70.82\% & 69.37\% & 71.69\% & 73.21\% \\

        \midrule
        \multicolumn{8}{c}{2-Shots}\\
        \midrule
        CUBirds & 83.82 & 85.82\% & 86.92\% & 86.12\% & 88.40\% & 86.77\% & 86.50\% \\
        DescribableTextures & 67.29 & 75.95\% & 76.57\% & 71.12\% & 78.99\% & 75.22\% & 76.17\% \\
        FGVCAircrafts & 47.44 & 52.30\% & 57.90\% & 55.15\% & 58.31\% & 55.82\% & 57.30\% \\
        Flowers-101 & 81.14 & 86.92\% & 85.05\% & 73.82\% & 61.25\% & 77.17\% & 79.50\% \\
        \midrule
        Avg & 69.92 & 75.25\% & 76.61\% & 71.56\% & 71.74\% & 73.75\% & 74.87\% \\
        
        \midrule
        \multicolumn{8}{c}{4-Shots}\\
        \midrule
        CUBirds & 83.82 & 88.30\% & 88.42\% & 86.72\% & 88.77\% & 87.50\% & 87.47\% \\
        DescribableTextures & 67.29 & 81.02\% & 77.90\% & 73.82\% & 78.85\% & 77.40\% & 79.82\% \\
        FGVCAircrafts & 47.44 & 58.65\% & 60.55\% & 55.37\% & 58.26\% & 57.55\% & 60.10\% \\
        Flowers-101 & 81.14 & 91.95\% & 87.07\% & 77.50\% & 65.44\% & 79.82\% & 81.42\% \\
        \midrule
        Avg & 69.92\% & 79.98\% & 78.49\% & 73.36\% & 72.83\% & 75.57\% & 77.21\% \\

        \midrule
        \multicolumn{8}{c}{8-Shots}\\
        \midrule
        CUBirds & 83.82 & 89.75\% & 88.40\% & 87.22\% & 88.68\% & 87.42\% & 87.32\% \\
        DescribableTextures & 67.29 & 85.07\% & 81.47\% & 75.82\% & 79.96\% & 79.90\% & 81.77\% \\
        FGVCAircrafts & 47.44 & 62.07\% & 61.55\% & 57.40\% & 58.56\% & 58.77\% & 61.72\% \\
        Flowers-101 & 81.14 & 94.30\% & 88.67\% & 79.47\% & 68.72\% & 80.32\% & 82.52\% \\
        \midrule
        Avg & 69.92\% & 82.80\% & 80.02\% & 74.98\% & 73.98\% & 76.61\% & 78.34\% \\

        \midrule
        \multicolumn{8}{c}{16-Shots}\\
        \midrule
        CUBirds & 83.82 & 90.32\% & 89.27\% & 89.45\% & 89.24\% & 87.97\% & 88.40\% \\
        DescribableTextures & 67.29 & 87.95\% & 83.20\% & 77.40\% & 81.12\% & 79.25\% & 81.50\% \\
        FGVCAircrafts & 47.44 & 65.57\% & 62.37\% & 57.85\% & 59.84\% & 61.02\% & 61.37\% \\
        Flowers-101 & 81.14 & 95.90\% & 90.15\% & 78.10\% & 71.23\% & 82.92\% & 84.90\% \\
        \midrule
        Avg & 69.92\% & 84.94\% & 81.25\% & 75.70\% & 75.36\% & 77.79\% & 79.04\% \\
        
    \bottomrule
    \end{tabular}
    \label{tab:steel_vision_accuracies}
\end{table}
\subsection{Attribution and Reliability of the Comparison}\label{app:steel_attribution}

\paragraph{Attribution.} To control for factors of variation in our experiments, we share the CLIP backbone and CoOp parameterization (1{,}024 tunable parameters). For evaluation episodes, we draw IID tasks per evaluation sample from the same class splits, 5-way, of varying number of shots. Due to random number generator variation between experiments, we can not guarantee the same episodes were sampled between every method. This sampling variation adds noise to comparisons between methods but no bias. Moreover, the multi-task methods, such as LoRA-Hub, MetaPB and \ourShort{}, all exploit the same upstream zoo (compute-matched for MetaPB; Appendix~\ref{app:steel_vision_hyperparams}). 

In terms of hyperparameters, as mentioned earlier, we use authors' default reported hyperparameters for LoRA-Hub and BBPT. For \ourShort{}, we use a set of fixed hyperparameters that were chosen at the beginning of the project on a set of toy experiments and never tuned further. For MetaPB, we tuned the hyperparameters qualitatively on meta-training distribution to achieve healthy training curves (meta-training loss is lowering and converging within 24 hours; the same GPU cost we give to \ourShort{}'s diffusion model training). This is covered in Appendix \ref{app:steel_vision_hyperparams}.

\paragraph{Reliability.} Table~\ref{tab:steel_vision_bound_stats_ste} reports accuracy with standard errors. For a valid comparison, we bootstrap over episodes by resampling each method's episodes independently with replacement (2K resamples), recompute each statistic and report the 95\% confidence interval of the difference (Table~\ref{tab:attribution_bootstrap}). \ourShort{}'s median bound is lower than the non-vacuous multi-task baselines on all four datasets, with every 95\% confidence interval excluding zero. Accuracy differences are resolved on Flowers-102 against both baselines and on Describable Textures against LoRA-Hub; elsewhere the intervals include zero, and we describe \ourShort{}'s accuracy as comparable. The difference between methods is the method of certification. For each method, we pair it with, to the best of our knowledge, the tightest published certificate that trains and certifies on the full support set.

\begin{table}[t]
    \centering
    \caption{16-shot 5-way: bound statistics with standard errors. Standard errors for the non-vacuous ratio, average gap and median bound are bootstrap estimates (2000 resamples with replacement); the accuracy standard error is the standard error of the mean. We report no standard error for a non-vacuous ratio of 0\% or 100\%, since the bootstrap estimate is zero by construction.}
    \small
    \resizebox{\columnwidth}{!}{
    \begin{tabular}{l>{\columncolor{Gray}}rrrrr}
        \toprule
        Method & SGD & BBPT & LoRA-Hub & MetaPB & \ourShort \\
        \midrule
        \multicolumn{6}{c}{\textbf{CUBirds}}\\
        \midrule
        Non-Vacuous Ratio & 0.00 & 0.00 & 100.00 & 100.00 & 100.00 \\
        Average Gap & 2.49 $\pm$ 0.01 & 2.48 $\pm$ 0.01 & 0.39 $\pm$ 0.01 & 0.29 $\pm$ 0.01 & 0.24 $\pm$ 0.01 \\
        Min Bound & 2.55 & 2.55 & 0.44 & 0.31 & 0.30 \\
        Median Bound & 2.58 $\pm$ 0.01 & 2.59 $\pm$ 0.01 & 0.50 $\pm$ 0.01 & 0.39 $\pm$ 0.01 & 0.36 $\pm$ 0.01 \\
        Max Bound & 2.64 & 2.68 & 0.61 & 0.54 & 0.47 \\
        Average Accuracy & 90.32 $\pm$ 0.78 & 89.27 $\pm$ 0.50 & 87.82 $\pm$ 0.84 & 89.56 $\pm$ 0.64 & 88.40 $\pm$ 0.72 \\

        \midrule
        \multicolumn{6}{c}{\textbf{Describable Textures}}\\
        \midrule        
        Non-Vacuous Ratio & 0.00 & 0.00 & 100.00 & 100.00 & 100.00 \\
        Average Gap & 2.43 $\pm$ 0.01 & 2.46 $\pm$ 0.01 & 0.39 $\pm$ 0.01 & 0.31 $\pm$ 0.01 & 0.24 $\pm$ 0.01 \\
        Min Bound & 2.55 & 2.55 & 0.48 & 0.40 & 0.36 \\
        Median Bound & 2.55 $\pm$ 0.00 & 2.63 $\pm$ 0.01 & 0.61 $\pm$ 0.01 & 0.51 $\pm$ 0.01 & 0.42 $\pm$ 0.01 \\
        Max Bound & 2.58 & 2.78 & 0.74 & 0.59 & 0.53 \\
        Average Accuracy & 87.95 $\pm$ 0.58 & 83.20 $\pm$ 0.76 & 77.60 $\pm$ 0.90 & 80.15 $\pm$ 0.72 & 81.50 $\pm$ 0.96 \\

        \midrule
        \multicolumn{6}{c}{\textbf{FGVCAircrafts}}\\
        \midrule        
        Non-Vacuous Ratio & 0.00 & 0.00 & 37.50 $\pm$ 7.71 & 85.00 $\pm$ 5.54 & 97.50 $\pm$ 2.43 \\
        Average Gap & 2.31 $\pm$ 0.01 & 2.45 $\pm$ 0.01 & 0.41 $\pm$ 0.01 & 0.33 $\pm$ 0.01 & 0.22 $\pm$ 0.01 \\
        Min Bound & 2.58 & 2.64 & 0.59 & 0.58 & 0.45 \\
        Median Bound & 2.64 $\pm$ 0.01 & 2.80 $\pm$ 0.02 & 0.83 $\pm$ 0.02 & 0.73 $\pm$ 0.02 & 0.61 $\pm$ 0.02 \\
        Max Bound & 2.78 & 3.01 & 1.05 & 0.95 & 0.85 \\
        Average Accuracy & 65.57 $\pm$ 1.45 & 62.37 $\pm$ 1.60 & 57.70 $\pm$ 1.67 & 59.63 $\pm$ 1.28 & 61.37 $\pm$ 1.25 \\
                
        \midrule
        \multicolumn{6}{c}{\textbf{Flowers-102}}\\
        \midrule
        Non-Vacuous Ratio & 0.00 & 0.00 & 87.50 $\pm$ 5.30 & 65.00 $\pm$ 7.63 & 100.00 \\
        Average Gap & 2.51 $\pm$ 0.00 & 2.50 $\pm$ 0.01 & 0.41 $\pm$ 0.01 & 0.49 $\pm$ 0.01 & 0.27 $\pm$ 0.01 \\
        Min Bound & 2.55 & 2.55 & 0.45 & 0.55 & 0.31 \\
        Median Bound & 2.55 $\pm$ 0.00 & 2.59 $\pm$ 0.01 & 0.61 $\pm$ 0.03 & 0.73 $\pm$ 0.03 & 0.40 $\pm$ 0.02 \\
        Max Bound & 2.55 & 2.70 & 1.03 & 0.95 & 0.61 \\ 
        Average Accuracy & 95.90 $\pm$ 0.46 & 90.15 $\pm$ 0.95 & 78.22 $\pm$ 2.30 & 72.02 $\pm$ 2.26 & 84.90 $\pm$ 1.43 \\
        
        \bottomrule
    \end{tabular}
    }
    \label{tab:steel_vision_bound_stats_ste}
\end{table}

\begin{table}[h]
\centering
\small
\caption{Bootstrap over episodes at 16 shots (episodes resampled independently): \ourShort{} minus baseline, with 95\% intervals. Negative bound differences favour \ourShort{}; positive accuracy differences (percentage points) favour \ourShort{}.}
\label{tab:attribution_bootstrap}
\begin{tabular}{llcc}
\toprule
Dataset & Baseline & $\Delta$ Median Bound & $\Delta$ Accuracy \\
\midrule
CUBirds              & LoRA-Hub & $-0.142$ $[-0.167, -0.117]$ & $+0.58$ $[-1.65, 2.73]$ \\
                     & MetaPB  & $-0.032$ $[-0.060, -0.009]$ & $-1.16$ $[-2.97, 0.77]$ \\
Describable Textures & LoRA-Hub & $-0.192$ $[-0.217, -0.154]$ & $+3.90$ $[1.37, 6.38]$ \\
                     & MetaPB  & $-0.086$ $[-0.111, -0.061]$ & $+1.35$ $[-0.90, 3.59]$ \\
FGVCAircrafts        & LoRA-Hub & $-0.223$ $[-0.273, -0.179]$ & $+3.68$ $[-0.43, 7.55]$ \\
                     & MetaPB  & $-0.124$ $[-0.173, -0.087]$ & $+1.75$ $[-1.79, 5.21]$ \\
Flowers-102          & LoRA-Hub & $-0.210$ $[-0.261, -0.142]$ & $+6.68$ $[1.50, 11.80]$ \\
                     & MetaPB  & $-0.336$ $[-0.403, -0.284]$ & $+12.88$ $[7.50, 18.24]$ \\
\bottomrule
\end{tabular}
\end{table}

\subsection{Vision Results with Standard Errors (STE)}\label{app:steel_vision_ste}
We report the performance of the 16-shot 5-way vision experiments in Table~\ref{tab:steel_vision_bound_stats_ste}. For the non-vacuous ratio, average gap, and median bounds, we compute the standard error using the bootstrap method. Specifically, we sample with replacement and report the standard deviation of the resulting distribution as the estimated standard error. For accuracy, computing STE is straightforward.

\begin{table}[t]
    \centering
    \caption{Aggregate over 16-Shots 5-way visual recognition episodes with Standard Errors}
    \small
    \begin{tabular}{l>{\columncolor{Gray}}rrrr}
        \toprule
        Method & SGD & BBPT & Meta-PB & \ourShort \\
        \midrule
        \multicolumn{4}{c}{\textbf{CUBirds}}\\
        \midrule
        Non-Vacuous Ratio & 0.00 ± 0.00 & 0.00 ± 0.00 & 97.50 ± 2.47 & 100.00 ± 0.00 \\
        Average Gap & 2.49 ± 0.01 & 2.48 ± 0.01 & 0.45 ± 0.01 & 0.24 ± 0.01 \\
        Min Bound & 2.55 & 2.55 & 0.34 & 0.30 \\
        Median Bound & 2.58 ± 0.01 & 2.59 ± 0.01 & 0.55 ± 0.01 & 0.36 ± 0.01 \\
        Max Bound & 2.64 & 2.68 & 0.85 & 0.47 \\
        Average Accuracy & 90.32 ± 0.78 & 89.27 ± 0.50 & 89.24 ± 0.70 & 88.40 ± 0.72 \\

        \midrule
        \multicolumn{4}{c}{\textbf{Describable Textures}}\\
        \midrule        
        Non-Vacuous Ratio & 0.00 ± 0.00 & 0.00 ± 0.00 & 50.00 ± 7.91 & 100.00 ± 0.00 \\
        Average Gap & 2.43 ± 0.01 & 2.46 ± 0.01 & 0.54 ± 0.01 & 0.24 ± 0.01 \\
        Min Bound & 2.55 & 2.55 & 0.57 & 0.36 \\
        Median Bound & 2.55 ± 0.00 & 2.63 ± 0.01 & 0.80 ± 0.02 & 0.42 ± 0.01 \\
        Max Bound & 2.58 & 2.78 & 1.06 & 0.53 \\
        Average Accuracy & 87.95 ± 0.58 & 83.20 ± 0.76 & 81.12 ± 0.68 & 81.50 ± 0.96 \\
        
        \midrule
        \multicolumn{4}{c}{\textbf{FGVCAircrafts}}\\
        \midrule        
        Non-Vacuous Ratio & 0.00 ± 0.00 & 0.00 ± 0.00 & 0.00 ± 0.00 & 97.50 ± 2.47 \\
        Average Gap & 2.31 ± 0.01 & 2.45 ± 0.01 & 0.86 ± 0.02 & 0.22 ± 0.01 \\
        Min Bound & 2.58 & 2.64 & 0.87 & 0.45 \\
        Median Bound & 2.64 ± 0.01 & 2.80 ± 0.02 & 1.25 ± 0.03 & 0.61 ± 0.02 \\
        Max Bound & 2.78 & 3.01 & 1.76 & 0.85 \\
        Average Accuracy & 65.57 ± 1.45 & 62.37 ± 1.60 & 59.84 ± 1.21 & 61.37 ± 1.25 \\
        
        \midrule
        \multicolumn{4}{c}{\textbf{Flowers-101}}\\
        \midrule
        Non-Vacuous Ratio & 0.00 ± 0.00 & 0.00 ± 0.00 & 10.00 ± 4.74 & 100.00 ± 0.00 \\
        Average Gap & 2.51 ± 0.00 & 2.50 ± 0.01 & 0.68 ± 0.03 & 0.27 ± 0.01 \\
        Min Bound & 2.55 & 2.55 & 0.65 & 0.31 \\
        Median Bound & 2.55 ± 0.00 & 2.59 ± 0.01 & 1.15 ± 0.10 & 0.40 ± 0.02 \\
        Max Bound & 2.55 & 2.70 & 1.77 & 0.61 \\
        Average Accuracy & 95.90 ± 0.46 & 90.15 ± 0.95 & 71.23 ± 2.40 & 84.90 ± 1.43 \\
        \bottomrule
    \end{tabular}
    \label{tab:steel_vision_bound_stats_ste}
\end{table}

\subsection{Error versus Complexity for All Methods}\label{app:steel_full_complexity}

\begin{figure}[t]
\centering
\includegraphics[width=\textwidth]{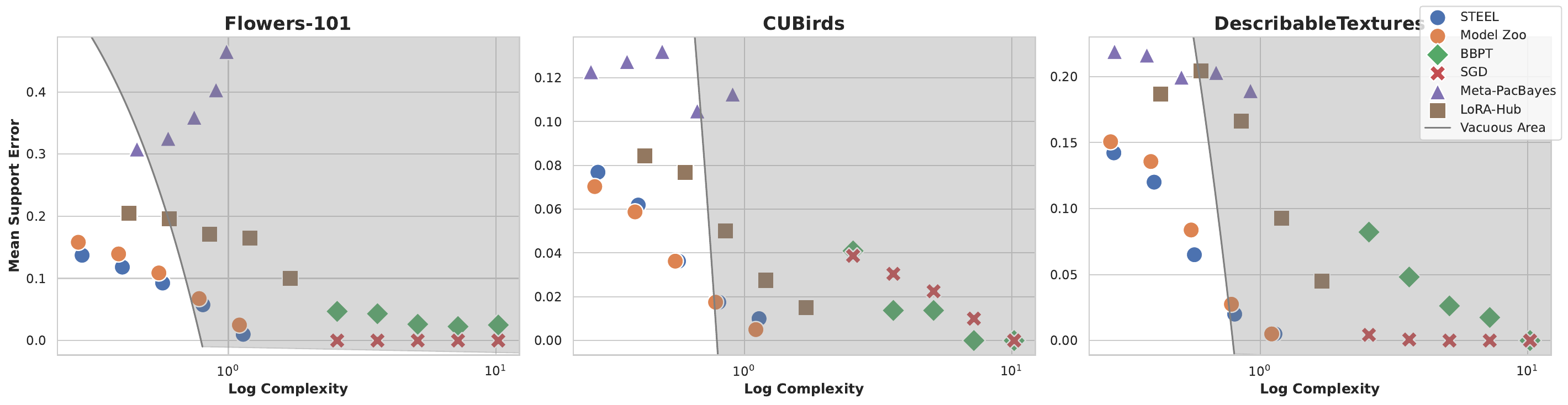}
\includegraphics[width=\textwidth]{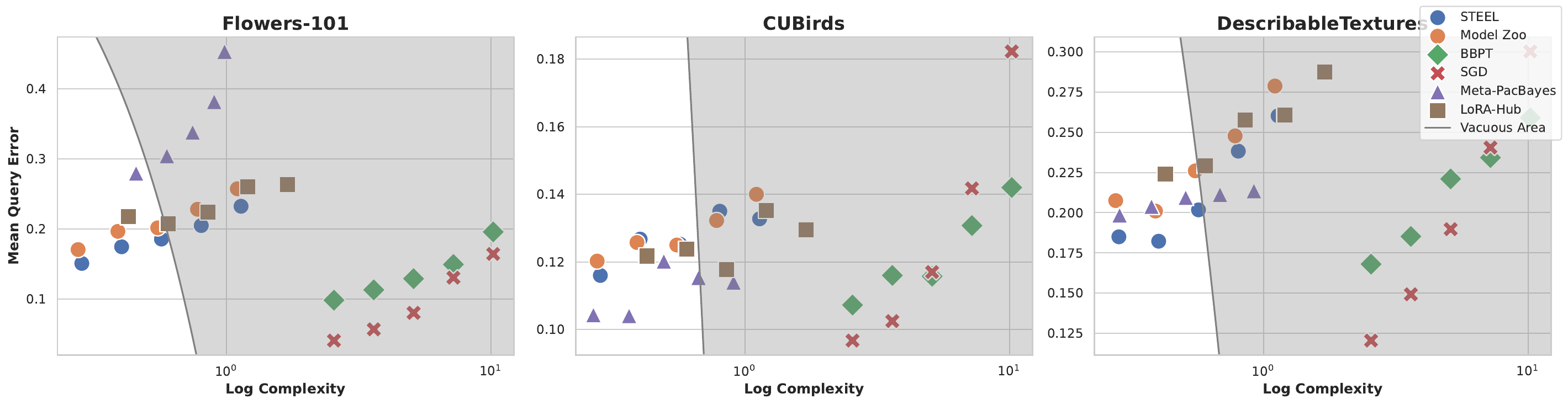}
\caption{Mean support (top) and query (bottom) error versus complexity term for all methods, 5-way, 1--16 shots per class. Shaded regions indicate vacuous bounds. Moving right to left, the number of shots increases.}
\label{fig:steel_full_complexity}
\end{figure}

Figure~\ref{fig:steel_full_complexity} extends Figure~\ref{fig:steel_teaser_clip} with the multi-task baselines LoRA-Hub and MetaPB. Each marker is one shot count; since the complexity term shrinks as $1/\sqrt{n}$, moving from right to left corresponds to increasing the number of shots from 1 to 16. For MetaPB, the y-axis is its Monte Carlo-corrected Gibbs support error and the x-axis its PAC-Bayes complexity term (Appendix~\ref{appsec:pb_meta}).

\paragraph{Support error.} The single-task learners fit the support set mostly better than multi-task baselines, but their complexity terms place every point in the vacuous region. For \ourShort{} and Model Zoo, support error rises with the number of shots: a finite hypothesis class cannot overfit a large support set. Please note on CUBirds, MetaPB attains a smaller complexity term than \ourShort{}, while \ourShort{} attains a lower support error; \ourShort{}'s tighter certificate comes from the sum of both support error and complexity term which makes up a certificate.

\paragraph{Query error.} The query panel is less orderly. SGD and BBPT reach lower query error for increasing shots, consistent with their higher accuracy in Table~\ref{tab:steel_vision_bound_stats}, but are never certified. Among the certifiable methods, MetaPB has the lowest mean query error on CUBirds, consistent with the statistically indistinguishable accuracy difference at 16 shots (Appendix~\ref{app:steel_attribution}). On the other datasets, \ourShort{}'s query error is comparable to or lower than that of the multi-task baselines.

\section{\ourShort{} Pseudo-Code}\label{app:steel_pseudo_code}
We provide pseudo-code in Algorithm~\ref{alg:transfer_learning}. There are upstream and downstream learning phases, as per meta-learning, model-zoo, and other transfer learning workflows. We refer to these as meta-training and meta-testing, to borrow terminology from the meta-learning literature ~\cite{hospedales20201metaSurveyPAMI,wang2019fewShotSurvey}.  The upstream meta-training phase trains one model per-task and then treats these models as data to learn a weight-space diffusion model. This effectively compresses upstream models into a single model generator that provides a prior finite hypothesis class. The downstream meta-testing phase performs low-shot learning by sampling from the generator and returning the model that minimises the downstream empirical risk. This leads strong generalisation guarantees in practice, because the finite hypothesis class incurs a low complexity cost in the certificate calculation.

\begin{algorithm}[t]
\caption{Transfer Learning via Diffusion and Finite Hypothesis Space}
\label{alg:transfer_learning}

\textbf{Meta-Training Phase:}
\begin{algorithmic}[1]
\STATE \textbf{Input:} Training tasks $\{T_1, T_2, \ldots, T_N\}$ sampled i.i.d. from $p_{\text{true}}(T)$
\STATE \textbf{Output:} Trained diffusion model $p(\theta)$

\FOR{$i = 1$ to $N$}
    \STATE Train PEFT adapter on task $T_i$ to obtain optimal parameters $\theta_i$
    \STATE $\overline{\Theta} \leftarrow \overline{\Theta} \cup \{\theta_i\}$ \COMMENT{Collect task-specific parameters}
\ENDFOR

\STATE Train diffusion model $p(\theta)$ using parameter set $\overline{\Theta} = \{\theta_1, \theta_2, \ldots, \theta_N\}$ as training data
\end{algorithmic}
\textbf{Meta-Testing Phase:}
\begin{algorithmic}[1]
\STATE \textbf{Input:} New task $T^* \sim p_{\text{true}}(T)$ with support set $S^* = \{(x_i, y_i)\}_{i=1}^n$
\STATE \textbf{Input:} Trained diffusion model $p(\theta)$, hypothesis space size $|\Theta|$
\STATE \textbf{Output:} Selected parameters $\theta^*$ and generalization bound

\STATE Sample finite hypothesis space: $\Theta = \{\theta^{(1)}, \theta^{(2)}, \ldots, \theta^{(|\Theta|)}\} \sim p(\theta)$

\STATE \textbf{Evaluate-then-Select Strategy:}\\
\COMMENT{Can use hierarchical search}
\FOR{$j = 1$ to $|\Theta|$}
    \STATE Compute empirical risk: $r(\theta^{(j)}) = \frac{1}{n} \sum_{(x,y) \in S^*} l(\theta^{(j)}; x, y)$
\ENDFOR

\STATE Select optimal parameters: $\theta^* = \arg\min_{\theta \in \Theta} r(\theta)$

\STATE \textbf{Compute Risk Certificate:}
\STATE With probability $\geq 1-\epsilon$, the generalization error satisfies:
\STATE $R(\theta^*) \leq r(\theta^*) + C \cdot \sqrt{\frac{\log\frac{|\Theta|}{\epsilon}}{2n}}$
\end{algorithmic}


\end{algorithm}

\section{Diffusion Model Ablation}\label{app:steel_diff_abl}

The use of a diffusion model in \ourShort{} is motivated by recent work on generative modeling in weight space~\citep{nndiffu}, which demonstrates that diffusion models can effectively capture the complex, multimodal distribution of neural network parameters. To evaluate whether this modeling capacity is critical for performance, we conduct an ablation study comparing against two alternative generative baselines for sampling parameter candidates in the LoRA-XS subspace:

\begin{itemize}
    \item \textbf{Random Sampling:} Each parameter is sampled independently from a Gaussian distribution whose mean and standard deviation are estimated from the corresponding statistics of the model zoo. This baseline assumes a fully factorized, unimodal distribution and requires no training or hyperparameter tuning.
    
    \item \textbf{Gaussian Mixture Model (GMM):} A GMM with diagonal covariance matrices is trained on the model zoo parameters. The number of mixture components is selected from $\{2, 4, 8, 16, 32, 64\}$ using the Bayesian Information Criterion (BIC). Once trained, the GMM is used to sample candidate parameter vectors.
\end{itemize}

\begin{table}[t]
    \centering
    \caption{Comparison of generative models for sampling parameter candidates on LaMP-2. The diffusion model is the only method that improves upon both the zero-shot baseline and SGD fine-tuning.}
    \begin{tabular}{lccccc}
        \toprule
        Method & Zero-Shot & SGD & Random & GMM & Diffusion \\
        \midrule
        Accuracy & 61.68\% & 63.25\% & 0.00\% & 60.29\% & \textbf{63.74\%} \\
        \bottomrule
    \end{tabular}
    \label{tab:steel_diff_abl}
\end{table}

The results in Table~\ref{tab:steel_diff_abl} demonstrate that the diffusion model significantly outperforms both random sampling and GMM-based sampling. Random sampling performs poorly, with zero-accuracy. Generated data from this baseline are random tokens. GMM improves over random but still fails to outperform the zero-shot baseline, likely due to its limited capacity to model high-dimensional parameter correlations and multimodality. In contrast, the diffusion model exceeds the performance of SGD-based fine-tuning, despite requiring no training on the test-time user data. This highlights its strong generative prior over parameter space.



\section{Robustness Under Distribution Shift}
\label{app:ood}

A natural question for any few-shot transfer learner is how it behaves when downstream tasks deviate from the upstream task distribution. Section~\ref{sec:bound_validity} established that the STEEL certificate (Eq.~\ref{eq:steel_bound}) remains \emph{valid} under distribution shift -- the i.i.d.\ assumption affects only \emph{tightness}, not correctness -- and Figure~\ref{fig:steel_inaturalist} demonstrated this on the CUB$\to$iNaturalist cross-domain transfer. Here we provide a more controlled study: (i) STEEL under graded, in-domain corruption (CUB-C), and (ii) a head-to-head comparison against MAML, the canonical gradient-based meta-learner whose inner-loop adaptation is often presumed to confer OOD robustness.

\subsection{Setup: CUB-C Corruption Benchmark}
\label{app:ood:setup}

Inspired by ImageNet-C~\citep{hendrycks2019benchmarking}, we construct \textbf{CUB-C} by applying three perturbations -- Gaussian noise, Gaussian blur, and brightness reduction -- to CUB images at three severity levels (mild, moderate, severe). This yields downstream tasks drawn from a distribution progressively further from the clean upstream tasks while remaining within the same semantic domain (fine-grained bird classification). All other meta-test settings match main text (16-shot, 5-way; CLIP backbone; CoOp prompt tuning; exhaustive search over $|\Theta|=20\text{K}$ STEEL samples). The diffusion model is the same one trained on clean CUB used throughout the main paper; no re-training is performed.

\subsection{STEEL Under Graded Distribution Shift}
\label{app:ood:steel_graded}

Table~\ref{tab:steel_cubc} reports STEEL's certificate behavior across severities. Three observations are salient.

\textbf{(1) Certificate validity is preserved.} STEEL maintains a $100\%$ non-vacuous ratio across all severity levels, including severe corruption that degrades empirical accuracy by $\sim$11.7 percentage points (88.60\%~$\to$~76.95\%). This empirically confirms the theoretical claim of Section~\ref{sec:bound_validity}: distribution shift inflates empirical risk $r(\theta)$, which loosens the right-hand side of Eq.~\ref{eq:steel_bound}, but does not invalidate the certificate.

\textbf{(2) The bound degrades gracefully.} The median bound rises from $0.35$ (clean) to $0.45$ (severe). The minimum bound is essentially flat ($0.28\to0.30$), reflecting the complexity term -- which depends only on $|\Theta|$ and the support size -- being unaffected by the shift. The maximum bound widens more substantially ($0.43\to0.80$), tracking the heaviest-tailed empirical-risk increases.

\textbf{(3) Certificates remain a useful deployment signal.} A practitioner inspecting a median bound of $0.45$ on severely corrupted data receives a principled, quantitative input to the deploy/no-deploy decision -- a signal that is structurally unavailable from uncertified methods, since empirical accuracy on a small support set is itself an unreliable estimator of test performance in the few-shot regime~\citep{shimabucoro2024evaluating}.

\begin{table}[t]
\centering
\caption{STEEL on CUB-C: certificate metrics across severity levels (16-shot, 5-way; 200 episodes). The non-vacuous ratio is preserved at 100\% across all severities, while the bound widens gracefully in proportion to empirical-risk degradation.}
\label{tab:steel_cubc}
\vspace{0.3em}
\small
\begin{tabular}{lcccc}
\toprule
& \textbf{Clean} & \textbf{Mild} & \textbf{Moderate} & \textbf{Severe} \\
\midrule
Non-Vacuous Ratio & 100\% & 100\% & 100\% & 100\% \\
Min Bound         & 0.28  & 0.30  & 0.30  & 0.30  \\
Median Bound      & 0.35  & 0.37  & 0.38  & 0.45  \\
Max Bound         & 0.43  & 0.52  & 0.61  & 0.80  \\
Avg.\ Accuracy    & 88.60\% & 86.70\% & 84.30\% & 76.95\% \\
\bottomrule
\end{tabular}
\end{table}

\subsection{Head-to-Head: STEEL vs.\ MAML Under Shift}
\label{app:ood:maml}

A common intuition is that gradient-based meta-learners such as MAML~\citep{maml} should outperform discrete-selection methods like STEEL under distribution shift, since the inner-loop SGD step in principle allows adaptation to move away from the meta-learned initialization. We test this directly by comparing the two methods on CUB-C under matched conditions (identical CLIP backbone, CoOp adapter, 16-shot 5-way episodes, same severity levels). MAML certificates are computed using the quantization bound of \citet{lotfi2024nonvacuousLLM} (Eq.~\ref{eq:qbound}), which is the tightest known bound applicable to its continuous hypothesis space.

\begin{table}[t]
\centering
\caption{MAML vs.\ STEEL on CUB-C (16-shot, 5-way). MAML provides zero non-vacuous certificates at every severity. Its accuracy advantage over STEEL is at most $\sim$1pp under mild shift, vanishes under moderate shift, and reverses under severe shift.}
\label{tab:maml_vs_steel}
\vspace{0.3em}
\small
\begin{tabular}{llcc}
\toprule
\textbf{Severity} & \textbf{Metric} & \textbf{MAML} & \textbf{STEEL} \\
\midrule
\multirow{3}{*}{Mild}
  & Non-Vacuous Ratio & 0\%   & \textbf{100\%}  \\
  & Median Bound      & 2.63  & \textbf{0.37}   \\
  & Avg.\ Accuracy    & \textbf{87.82\%} & 86.70\% \\
\midrule
\multirow{3}{*}{Moderate}
  & Non-Vacuous Ratio & 0\%   & \textbf{100\%}  \\
  & Median Bound      & 2.65  & \textbf{0.38}   \\
  & Avg.\ Accuracy    & \textbf{84.72\%} & 84.30\% \\
\midrule
\multirow{3}{*}{Severe}
  & Non-Vacuous Ratio & 0\%   & \textbf{100\%}  \\
  & Median Bound      & 2.69  & \textbf{0.45}   \\
  & Avg.\ Accuracy    & 76.47\% & \textbf{76.95\%} \\
\bottomrule
\end{tabular}
\end{table}

Two findings stand out (Table~\ref{tab:maml_vs_steel}).

\textbf{MAML provides no certificate at any severity.} Its median bound remains in the $2.6$--$2.7$ range across all conditions, well above the vacuous threshold. The continuous hypothesis space of MAML's adapted weights incurs a complexity term that small-$n$ certificates cannot overcome, irrespective of the empirical accuracy attained.

\textbf{MAML's empirical accuracy advantage is negligible and non-monotonic in shift severity.} The gap is $\sim$1pp at mild shift, effectively zero at moderate shift, and \emph{reverses} at severe shift (MAML $76.47\%$ vs.\ STEEL $76.95\%$). This directly contradicts the hypothesis that inner-loop SGD confers substantial OOD adaptation. MAML's adaptation is constrained to a local neighborhood of the meta-learned initialization -- whether by a small number of inner steps or by an explicit regularizer (e.g., iMAML~\citep{NEURIPS2019_072b030b}) -- and without such locality, adaptation reduces to plain SGD on a very small support set, which generalizes poorly. Neither approach \emph{fully} addresses OOD; STEEL's discrete selection delivers comparable accuracy while preserving certifiability.

\subsection{Scope and Limitations}
\label{app:ood:limits}

We deliberately scope this study to in-domain corruption rather than cross-dataset transfer (e.g., miniImageNet$\to$CUB). The CUB-C setup isolates the effect of input-distribution shift while holding the semantic task distribution fixed, which is the cleanest setting for assessing the \emph{validity} and \emph{graceful degradation} of the certificate. Cross-dataset OOD additionally entails task-distribution shift and is known to be challenging for the entire transfer/meta-learning paradigm; it also requires retraining the upstream diffusion model on a different source, which is outside the scope of our current study. We discuss this honestly as a limitation and outline concrete extensions of STEEL to this setting in Section~\ref{sec:discussion}, including: (i) conditioning the diffusion model on support-set embeddings to steer generation toward downstream-relevant regions while preserving the finite hypothesis space; (ii) hybrid hypothesis classes that union diffusion models trained on multiple source domains; and (iii) combining STEEL body selection with a lightweight gradient-based classifier head, at small additional certificate cost.

\subsection{Summary}
\label{app:ood:summary}

The certificate guarantee underlying STEEL is robust to in-domain distribution shift in the operational sense that matters for deployment: certificates remain valid, degrade gracefully with empirical risk, and provide a quantitative signal that uncertified methods structurally cannot. Against MAML -- the prototypical gradient-based meta-learner -- STEEL matches or exceeds empirical accuracy under moderate-to-severe shift while providing the only non-vacuous certificate of the two methods. Together with the CUB$\to$iNaturalist result of Figure~\ref{fig:steel_inaturalist}, these experiments substantiate the claim that STEEL's certifiability is not purely an in-distribution phenomenon.


\section{Limitations and Future Directions}
\label{sec:discussion}

STEEL establishes the feasibility of certifiable few-shot transfer learning with large models. Like any new paradigm, it leaves open a research program for tightening guarantees, closing the empirical-accuracy gap with continuous-hypothesis methods, and extending coverage beyond the in-distribution setting. We discuss these directions in three groups: tightening the certificate (Sec.~\ref{app:future:tighten}), improving empirical accuracy (Sec.~\ref{app:future:accuracy}), and extending STEEL to cross-domain and out-of-distribution (OOD) settings (Sec.~\ref{app:future:ood}). All directions preserve the structural property that gives STEEL its guarantee -- a tractable hypothesis space -- and are therefore drop-in modifications rather than rethinks of the framework.

\subsection{Tightening the Certificate}
\label{app:future:tighten}

\textbf{Covering-number refinement of the hypothesis space.} The bound in Eq.~\ref{eq:steel_bound} treats every sampled adapter as contributing independently to the complexity term $\log|\Theta|$. In practice, many sampled adapters produce near-identical predictions and should not both count toward complexity. What matters for generalisation is not $|\Theta|$ but the \emph{covering number} of $\Theta$ in prediction space: the minimum number of $\varepsilon$-balls needed to cover the induced predictor set under a suitable functional distance. A clean and immediate route to exploiting this within our existing bound is to pre-process $\Theta$ into an $\varepsilon$-net $\Theta' \subseteq \Theta$, obtained by clustering predictions on a held-out upstream reference set and retaining one representative per cluster. Running STEEL on $\Theta'$ leaves Eq.~\ref{eq:steel_bound} unchanged but replaces $\log|\Theta|$ with $\log|\Theta'| = \log\mathcal{N}(\Theta,\varepsilon)$. Because the reference set is upstream data, no dependence on the downstream support set $S^*$ is introduced, preserving the assumptions of the bound. Practitioners can then draw \emph{more} diffusion samples (improving empirical fit) while the effective hypothesis-space size grows only with prediction-space diversity rather than raw sample count. Pushing this further would require switching bound technology -- e.g., chaining or fat-shattering arguments -- which is a more ambitious extension.

\textbf{Data-dependent priors over $\Theta$.} Our certificate uses a uniform prior over the sampled set $\Theta$, which yields the $\log|\Theta|$ complexity term in Eq.~\ref{eq:steel_bound}. A tighter alternative is to use the diffusion model's own likelihood as a non-uniform prior: $\pi(\theta) \propto p_{\text{diff}}(\theta)$ for $\theta \in \Theta$. The PAC-Bayes complexity term then becomes $\mathrm{KL}(\delta_{\theta^*} \| \pi) = -\log \pi(\theta^*)$, which replaces $\log|\Theta|$ with $-\log p_{\text{diff}}(\theta^*) + \log Z$ (with $Z = \sum_{\theta \in \Theta} p_{\text{diff}}(\theta)$). Whenever the selected $\theta^*$ lies in a high-likelihood region of the diffusion model -- which is precisely what we expect when the upstream and downstream task distributions are aligned -- this term is smaller than $\log|\Theta|$ and the bound tightens. Because the diffusion model is trained on a disjoint upstream split, $\pi$ is independent of the downstream support set $S^*$, preserving the assumptions of the bound (cf.~\citet{ortiz2021tighter}). A practical consideration is that exact diffusion log-densities are intractable; however, the standard DDPM ELBO is a \emph{lower} bound on $\log p_{\text{diff}}(\theta^*)$, equivalently an \emph{upper} bound on $-\log p_{\text{diff}}(\theta^*)$, and substituting it into the certificate yields a valid (if slightly conservative) bound. The normalising constant $Z$ can likewise be upper-bounded using upstream samples. This direction repurposes information that STEEL already computes -- the diffusion likelihood -- to tighten the certificate without changing the algorithm.
\subsection{Improving Empirical Accuracy}
\label{app:future:accuracy}

\textbf{Stronger weight-space architectures.} STEEL's certificate validity depends only on the empirical risk of sampled adapters, not on the architecture of the generator. Recent weight-space learning methods such as SANE~\citep{10.5555/3692070.3693860} model neural-network parameter distributions substantially better than the MLP-based diffusion model used in our experiments. Substituting such architectures into the upstream phase is a drop-in improvement to sample quality -- and therefore to downstream accuracy -- without any change to the theoretical framework.

\textbf{Better PEFT modules.} Our experiments fix the PEFT method per backbone (LoRA-XS~\citep{baazy2024loraxs} for LLMs; CoOp~\citep{zhou2022coop} for vision). The choice of PEFT method affects both individual adapter quality and the structure of the parameter distribution the diffusion model must learn. Unified PEFT benchmarks~\citep{vpetl,Mai_2025_CVPR} document significant method-dependent variation in accuracy. Since lower adapter quality translates directly to higher $r(\theta)$ and looser bounds, swapping in stronger PEFT families as the landscape evolves is a practical lever for improving both accuracy and certificate tightness simultaneously.

\textbf{Mixture hypothesis classes across PEFT families.} Currently, a single diffusion model is trained over a single PEFT family. A natural extension is to train \emph{separate} diffusion models over different PEFT families and construct the downstream hypothesis space as their union. The complexity term increases only by a small constant ($\log K$ for $K$ families), while the hypothesis space gains substantially richer coverage. At meta-test time, STEEL automatically selects whichever PEFT family best fits the support set, providing implicit task-conditional PEFT selection at no asymptotic cost in certificate strength.

\subsection{Cross-Domain and OOD Extensions}
\label{app:future:ood}

Appendix~\ref{app:ood} shows that STEEL's certificate behaves well under graded in-domain corruption (CUB-C). Cross-\emph{dataset} transfer -- where upstream and downstream tasks differ in both input statistics and label semantics -- is more challenging for the entire transfer/meta-learning paradigm. We discuss three concrete directions for extending STEEL to this regime, each of which preserves certifiability.

\textbf{Scaling meta-training across diverse domains.} The cross-domain few-shot learning literature has converged on a consistent finding: robustness to domain shift is driven primarily by the breadth of the meta-training distribution rather than by the choice of meta-test adaptation procedure. Methods such as URL~\citep{url} and tri-M~\citep{trim} achieve strong cross-domain performance using \emph{frozen} or near-frozen test-time adaptation, succeeding because their upstream phase aggregates over many domains. Recent comparisons of single- vs.\ multi-source meta-dataset training~\citep{eustratiadis2024neural} confirm this picture. STEEL is structurally well-suited to multi-domain upstream training: because per-task adapters are trained independently before being aggregated into the diffusion model, scaling to additional domains amounts to extending the upstream pool. There is no need for the careful multi-domain episode balancing required by gradient-based meta-learners. The diffusion model can then learn a multi-modal distribution with modes per source dataset, and at meta-test it automatically selects samples in the vicinity of the most-related source mode.

\textbf{Support-set-conditional generation.} STEEL currently samples from the unconditional prior $p(\theta)$. Conditioning the diffusion model on a representation of the support set $S^*$ -- e.g., a permutation-invariant encoding of $\{(x_i, y_i)\}_{i=1}^n$ -- would steer generation toward downstream-relevant regions of weight space. This is the generative-model analogue of MAML's gradient-based adaptation: it uses the support set to inform \emph{where} to search, but does so within a finite hypothesis space sampled from $p(\theta \mid S^*)$. Certifiability is preserved (the hypothesis space remains finite), with a modest bound loosening to account for the conditioning channel that can be analysed via standard PAC-Bayes conditioning arguments.

\textbf{Sample-then-refine: hybrid discrete--continuous adaptation.} A natural way to combine STEEL's certifiability with the local adaptability of gradient-based methods is to select $\theta^*$ from $\Theta$ as in STEEL, and then apply bounded SGD updates within an $\varepsilon$-ball around $\theta^*$. The resulting hypothesis space is a finite union of bounded continuous regions, for which PAC-Bayes bounds with data-dependent priors remain applicable. A particularly tractable instantiation is the SimpleShot-style decomposition~\citep{simpleshot}: select the backbone/body via STEEL (preserving the discrete certificate over body parameters), then fit a linear classifier head with SGD on the support set. The classifier head admits a tight separate bound -- linear classifiers in low-shot regimes are exactly the setting where classical generalisation theory is sharp -- and the two bounds can be combined.

\subsection{A Research Programme}
\label{app:future:programme}

We close by situating STEEL in the broader trajectory of meta-learning research. Gradient-based meta-learning evolved incrementally over nearly a decade: MAML~\citep{maml} (2017) was demonstrated on single-domain benchmarks; Meta-Dataset~\citep{metadataset} (2020) introduced cross-domain evaluation; methods specifically designed for cross-domain transfer (URL, tri-M, TSA) followed in 2021--2022; and scaling laws for meta-learning have only recently come into view. STEEL introduces a fundamentally different paradigm -- discrete hypothesis selection from a learned weight-space distribution -- and establishes its feasibility together with the first non-vacuous certificates for few-shot deep learning. The directions above constitute a concrete and incremental research programme for extending this foundation: tighter certificates via prediction-space covering numbers and data-dependent priors; closing the empirical-accuracy gap via stronger weight-space architectures, better PEFT families, and mixture hypothesis classes; and reaching the cross-domain regime via multi-domain upstream scaling, conditional generation, and hybrid discrete--continuous adaptation. Establishing scaling laws for weight-space generative models -- analogous to those now well-understood for language models -- is a final, longer-horizon item that would let practitioners predict the certificate--accuracy frontier as a function of upstream compute, hypothesis-space size, and downstream support-set size.